\documentclass[10pt,twocolumn,letterpaper]{article}

\usepackage[pagenumbers]{cvpr} 

\usepackage{multirow}

\definecolor{cvprblue}{rgb}{0.21,0.49,0.74}
\usepackage[pagebackref,breaklinks,colorlinks,allcolors=cvprblue]{hyperref}
\usepackage{microtype}
\definecolor{teal}{rgb}{0, 0.5, 0.4}

\definecolor{st}{rgb}{0.9, 0.4, 0.0}

\newcommand\blfootnote[1]{
    \begingroup
    \renewcommand\thefootnote{}\footnote{#1} 
    \addtocounter{footnote}{-1}
    \endgroup
}

\title{Generative Video Propagation}

\author{
    Shaoteng Liu$^{1,*}$, Tianyu Wang$^2$, Jui-Hsien Wang$^2$, Qing Liu$^2$, Zhifei Zhang$^2$, \\
   \vspace{2mm}
    Joon-Young Lee$^2$, Yijun Li$^2$, Bei Yu$^1$, Zhe Lin$^2$, Soo Ye Kim$^{2,\dagger}$, Jiaya Jia$^{3,4,\dagger}$ \\
    \vspace{1mm}
   $^1$ The Chinese University of Hong Kong \quad $^2$ Adobe Research \\
   $^3$ The Hong Kong University of Science and Technology \quad $^4$ SmartMore \\
    \url{https://genprop.github.io//}
}

\begin{document}

\twocolumn[{
\renewcommand\twocolumn[1][]{#1}
\maketitle
\begin{center}
    \centering
    \captionsetup{type=figure}
    \includegraphics[width=0.96\textwidth]{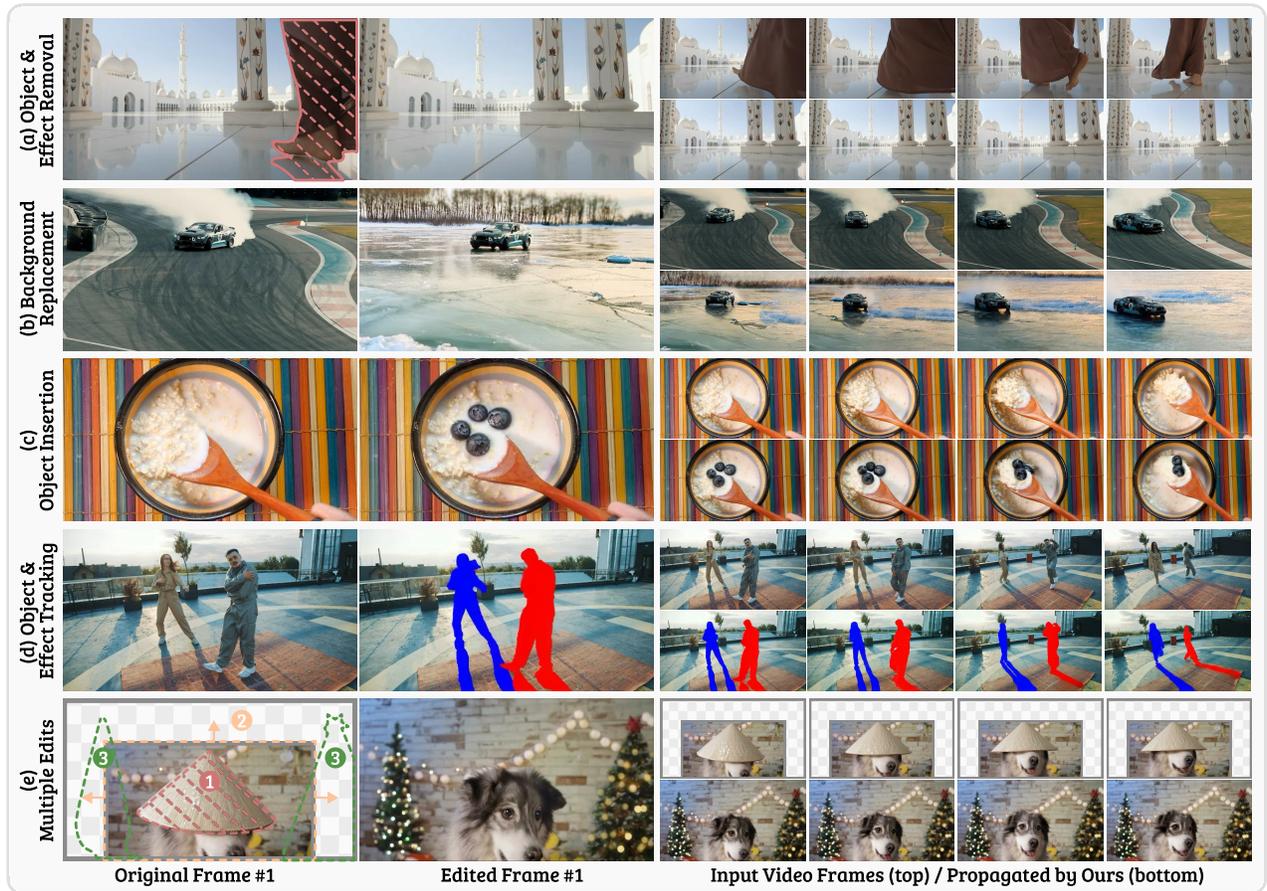}
    \captionof{figure}{
    GenProp. We propose a generative video propagation framework (GenProp), which can seamlessly propagate any first frame edit through the video. GenProp supports a wide range of video applications, including (a) complete object removal with effects such as shadows and reflections, (b) background replacement with realistic effects, (c) object insertion where inserted objects have physically plausible motion (i.e., blueberries falling while spoon goes up), (d) tracking of objects and their associated effects, and (e) multiple edits (outpainting, insertion, removal) at a single inference run.
    }
    \label{fig:teaser}
\end{center}
}]

\blfootnote{* Work done during an internship at Adobe.}
\blfootnote{$\dagger$ Co-corresponding authors.}

\begin{abstract}
Large-scale video generation models have the inherent ability to realistically model natural scenes. In this paper, we demonstrate that through a careful design of a \textbf{generative video propagation framework}, various video tasks can be addressed in a unified way by leveraging the generative power of such models. Specifically, our framework, GenProp, encodes the original video with a selective content encoder and propagates the changes made to the first frame using an image-to-video generation model. We propose a data generation scheme to cover multiple video tasks based on instance-level video segmentation datasets. Our model is trained by incorporating a mask prediction decoder head and optimizing a region-aware loss to aid the encoder to preserve the original content while the generation model propagates the modified region. This novel design opens up new possibilities: In editing scenarios, GenProp allows substantial changes to an object’s shape; for insertion, the inserted objects can exhibit independent motion; for removal, GenProp effectively removes effects like shadows and reflections from the whole video; for tracking, GenProp is capable of tracking objects and their associated effects together. Experiment results demonstrate the leading performance of our model in various video tasks, and we further provide in-depth analyses of the proposed framework.
\end{abstract}
\section{Introduction}
\label{sec:intro}

Recently, large-scale video generation models~\cite{sora,moviegen,hong2022cogvideo,yang2024cogvideox,chen2024videocrafter2,singer2022make,villegas2022phenaki,ho2022imagen} have shown impressive performance, generating highly realistic videos while being able to simulate the complexities of the real world. In this rapidly evolving domain, following works in video generation have extended the text-to-video (T2V) generation to image-to-video (I2V)~\cite{sora,pika,blattmann2023stable,zhang2023i2vgen,xing2025dynamicrafter}, and are further exploring various video editing tasks such as video inpainting~\cite{zi2024cococo}, appearance editing~\cite{ouyang2024i2vedit,singer2025video}, object insertion~\cite{mou2024revideo}, usually focusing on that specific task. In this paper, we bring a different perspective by observing that many of such video applications can be modeled as a more general \textit{video propagation} problem.

Video propagation itself is not a new concept, with traditional methods often relying on optical flow~\cite{teed2020raft,cong2023flatten},
depth~\cite{yan2023magicprop,ceylan2023pix2video}, radiance fields~\cite{ouyang2024codef}, and atlases~\cite{kasten2021layered,huang2023inve} to propagate the changes in sparse intermediate frames (typically the first frame) to the rest of the video. However, such approaches can be prone to error accumulation, leading to limited robustness and generalization ability. Furthermore, they often focus on a single task~\cite{cong2023flatten,ouyang2024codef} or entail retraining for a specific task for propagation~\cite{ouyang2024i2vedit,mou2024revideo,li2024vidtome,zhang2024towards}. In contrast, we define a new problem of \textit{generative video propagation} by leveraging the inherent power of video generation models in modeling real-world scenes.

Our model, \textit{GenProp}, is able to propagate the changes in the first frame to the whole video while keeping other parts consistent to the original video, without requiring any additional motion predictions (e.g., optical flow). This general formulation enables many downstream applications such as removal, insertion, replacement (object and/or background), text-based editing, outpainting and even object tracking, some of which are shown in Fig.~\ref{fig:teaser}. We further demonstrate that our model is also able to \textit{expand} the scope of what is usually achievable in each task, specifically: (1) substantial shape modifications in object editing tasks, (2) independent motion of inserted objects in insertion tasks, (3) removal of object effects like shadows and reflections in removal tasks, and (4) accurate tracking of objects along with their associated effects. Note that unlike existing video editing models that often require a dense mask labeling for all individual frames (e.g., for object removal), GenProp does not require any mask input, thanks to the propagation-based approach, greatly simplifying the editing process.

\begin{figure}
  \centering
   \includegraphics[width=1\linewidth]{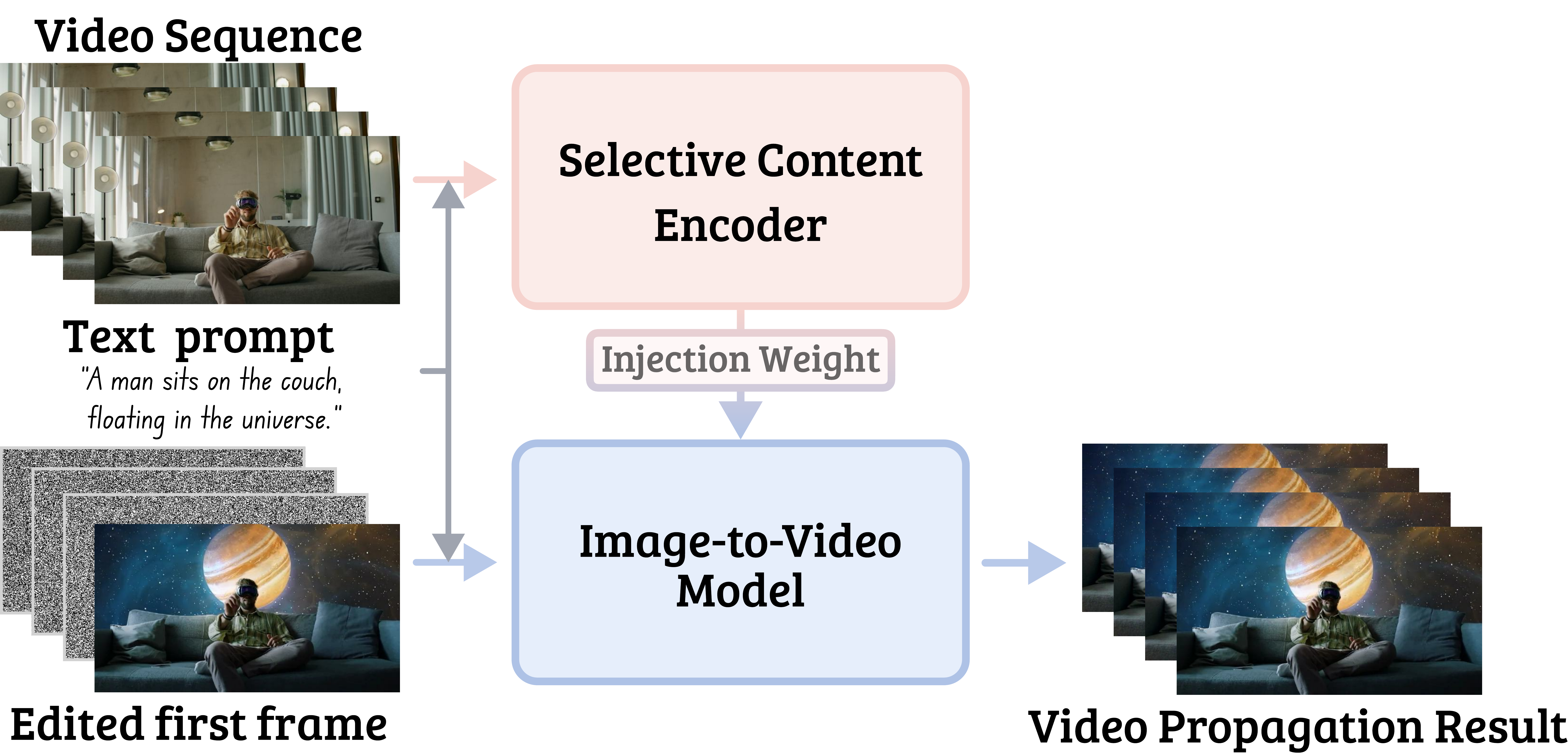}

   \caption{
   Model Overview.
   During inference, our framework takes in the original video as input through a selective content encoder (SCE) to retain content in unchanged regions.
   Changes applied to the first frame are propagated throughout the video using an I2V model while other regions remain intact.}
   \vspace{-2mm}
   \label{fig:framework-general}
\end{figure}

\begin{figure}
  \centering
   \includegraphics[width=1\linewidth]{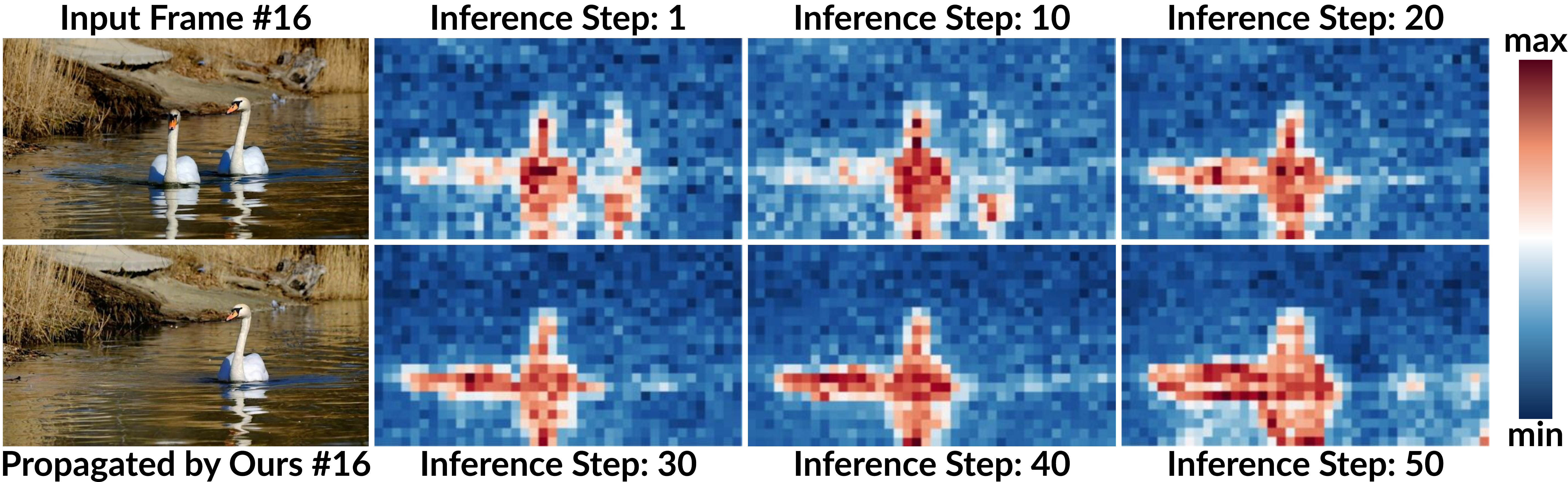}

   \caption{Attention Map Visualization. We observe that the attention maps gradually focus on the regions to be removed and the I2V model is guided to generate new content in those regions.}
   \vspace{-2mm}
   \label{fig:attn-vis}
\end{figure}

Our model architecture consists of two main components as shown in Fig.~\ref{fig:framework-general}: the Selective Content Encoder (SCE) that encodes the information of the original video, and the I2V generation model that takes in the edited first frame for propagation. The training objective is to allow SCE to selectively encode the features of the unchanged parts of the video, while preserving the generation capabilities of I2V models to propagate the altered parts. To effectively disentangle these two functions, we introduce a region-aware loss and penalize the gradients within the modified region for SCE, as ideally, SCE should not encode content in the edited area. 
For training the model, we propose using synthetic data derived from video instance segmentation datasets.
As shown in the attention map visualizations in Fig~\ref{fig:attn-vis}, we observe that GenProp indeed attends to the region to be modified and the I2V model is guided to generate (propagate) the new content into those regions. To further aid the model, we incorporate an auxiliary decoder head during training to predict the modified region.

Our contributions are summarized as follows:
\begin{itemize}
\item We define a novel problem of \textit{generative video propagation} that aims to propagate various changes in the first frame of the video to the entire video by leveraging the generative power of I2V models.
\item We carefully design our model, \textit{GenProp}, with a Selective Content Encoder (SCE), dedicated loss functions and a mask prediction head and propose a synthetic data generation pipeline for training this model.
\item Our model supports various downstream applications such as removal, insertion, replacement, editing, and tracking. We observe that it further supports outpainting even without any task-specific data during training. 
\item Experiment results show that our model outperforms SOTA methods in video editing and object removal while expanding the scope of existing tasks including tracking.
\end{itemize}
\section{Related Work}
\label{sec:rw}

\noindent\textbf{Video Propagation.} Traditional methods are typically designed for a single task and often require retraining for new tasks~\cite{jampani2017video,cong2023flatten,ouyang2024codef}. Many approaches address propagation by first tracking instance masks, then performing inpainting~\cite{kim2019deep,ke2021occlusion}, with segmentation often as the initial step. SAM 2~\cite{ravi2024sam2}, the current SotA tracking model, can track the masks accurately and efficiently. Some methods rely on optical flow~\cite{teed2020raft,cong2023flatten} or depth~\cite{yan2023magicprop,ceylan2023pix2video} to ensure consistent motion and spatial coherence. CoDef~\cite{ouyang2024codef} uses deformation fields from the source video to guide edits from the first frame. While these representations aid motion tracking and structural consistency, they add complexity and may limit flexibility, especially with significant shape changes or complex backgrounds. Depth, sketches, and optical flow can also be combined with diffusion models~\cite{esser2023structure,liang2024flowvid,xing2024make,wang2024videocomposer,yan2023motion}.

\vspace{0.5em}
\noindent\textbf{Diffusion-based Video Editing.} Most diffusion-based video editing methods rely on text control, where the primary goal is to make edits that are coherent to text prompts while preserving the unchanged regions of the video. Some methods utilize text-to-image models for zero-shot editing through attention control~\cite{qi2023fatezero,ceylan2023pix2video,geyer2023tokenflow,kara2024rave,wang2023zero,wu2024fairy,shin2024edit}. Some other works require intermediate variables like optical flows or depth maps to stabilize motion. Others rely on per-case fine-tuning to adapt to specific motion~\cite{wu2023tune,liu2024video,bar2022text2live,zhao2023controlvideo}, but this approach is typically slow and prone to generate similar results from the original video due to reconstruction tuning. SORA~\cite{sora} denoises the noised videos under the target description to do editing.
These methods are generally limited to altering the appearance rather than making significant changes to object shapes. Additionally, because of unclear attention maps, especially in complex scenes, background changes often lack precision and coherence.
InsV2V~\cite{cheng2023consistent} and EVE~\cite{singer2025video} edit videos based on text instructions but are also limited to appearance changes.
Some recent efforts have attempted to directly edit motion based on text prompts~\cite{jeong2024vmc,zhao2025motiondirector,yatim2024space}, but their resulting video output tends to strike a balance between the text-based guidance and the original video’s motion, which can be hard to control.

\vspace{0.5em}
\noindent\textbf{Image-to-Video Generation and Editing.} Image-to-video (I2V) generation models take an input image along with a text prompt to generate a sequence of frames, making them a foundational application in video generation due to
its familiarity and versatility.
Notable open-source models include Stable Video Diffusion~\cite{blattmann2023stable}, I2VgenXL~\cite{zhang2023i2vgen}, and SparseCtrl~\cite{guo2025sparsectrl}, while high-performance commercial models, such as Gen-2, PikaLabs~\cite{pika}, SORA~\cite{sora}, and Movie Gen~\cite{moviegen}, further push the boundaries in this field.

Several methods propagate edits based on modifications made to the first frame. For example, some works~\cite{shi2024motion,yan2023motion,yan2023magicprop,feng2024ccedit}
rely on first-frame edits but require auxiliary inputs like optical flow or depth maps for motion continuity. VideoSwap~\cite{gu2024videoswap} uses sparse key points to control the motion. AnyV2V~\cite{ku2024anyv2v} can also propagate first-frame edits across a video sequence; however, as a training-free framework, its generalization ability is limited. I2VEdit~\cite{ouyang2024i2vedit}, in contrast, necessitates learning motion LoRAs~\cite{hu2021lora} for each video clip, adding computational complexity.
Revideo~\cite{mou2024revideo}, built on Stable Video Diffusion (SVD), enables control over the generation using the edited first frame and a specified motion trajectory. However, its approach involves masking parts of the input video with a black square, which removes significant information and restricts the method in handling complex background edits and large shape alterations.

\section{Method}
\label{sec:method}

\begin{figure*}
  \centering
   \includegraphics[width=0.95\linewidth]{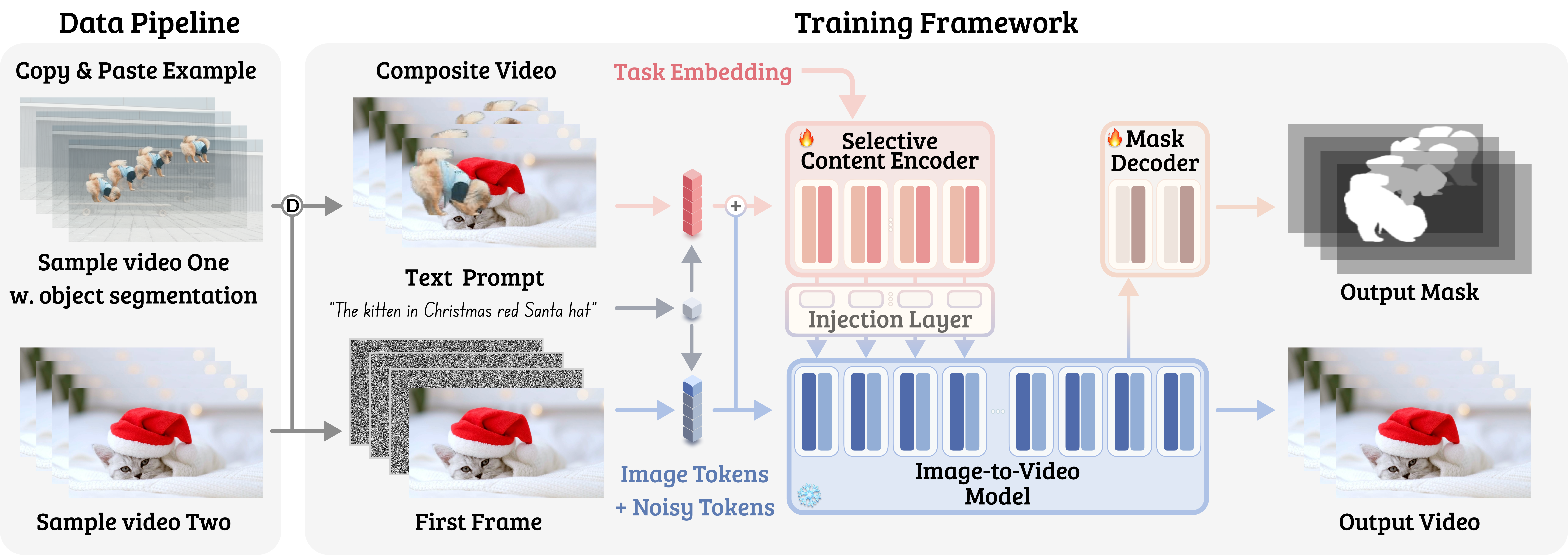}

   \caption{Training Framework of GenProp. Our framework integrates a Selective Content Encoder and a Mask Prediction Decoder on top of the I2V generation model, enforcing the model to propagate the edited region while preserving the content in the original video for all other regions. With synthetic data augmentations and task embeddings, our model is trained to propagate various changes in the first frame.}
   \vspace{-2mm}
   \label{fig:framework-detail}
\end{figure*}

Generative video propagation has the following key challenges: (1) Realism -- changes in the first frame should be naturally propagated to the following frames, (2) Consistency -- all other regions should remain consistent to the original video, and (3) Generality -- the model should be general enough to be applicable to multiple video tasks. In GenProp, we leverage an I2V generation model for (1); we introduce a selective content encoder and a mask prediction decoder and train the model with a region-aware loss to address (2); and we propose a data generation scheme and also benefit from the versatile I2V model for (3).

\subsection{Problem Formulation}
\label{sec:problem}

Given an input video \( V = \{v_1, v_2, \dots, v_T\} \) with \( T \) frames, let \( v'_1 \) denote the modified first frame. The goal is to propagate this modification, producing a modified video \( V' = \{v'_1, v'_2, \dots, v'_T\} \), where each frame \( v'_t \) (for \( t = 2, \dots, T \)) retains the modification applied to the key frame \( v_1 \) while maintaining consistency in both appearance and motion throughout the sequence.
We employ a latent diffusion model that encodes pixel information in the latent space. With a slight abuse of notations, we continue using $v_t$ for this latent representation.
In formal terms, during inference, GenProp generates each frame \( v'_t \) as:
\begin{equation}
v'_t = \mathcal{G}(\mathcal{E}(V), v'_1, t), \quad \forall t \in \{2, \dots, T\},
\end{equation}
where $\mathcal{G}$ is the I2V generation model guided by the selective content encoder (SCE), $\mathcal{E}(V)$.

For training, we use synthetic data constructed from existing video instance segmentation datasets to create paired samples (details given in Sec.~\ref{sec:data_gen}). We define a data generation operator \( \mathcal{D} \) that constructs training data pairs \( (v_i, \hat{v}_i) \) from an original video sequence \( V \).
Let \( \mathcal{D}(V) \) denote the synthetic data generation operator applied to the original video sequence, where:
\begin{equation}
(v_i, \hat{v}_i) \in \mathcal{D}(V), \quad \forall i \in \{1, \dots, T\}.
\end{equation}
Then \( \hat{V} = \{\hat{v}_1, \hat{v}_2, \dots, \hat{v}_T\} \) is the synthetic video sequence. GenProp is trained to satisfy the following objective across all frames \( i \in \{2, \dots, T\} \):
\begin{equation}
\min_{\mathcal{E}} \sum_{i=2}^{T} \mathcal{L}(\mathcal{G}(\mathcal{E}(\hat{V}), v_1, i), v_i)
\label{eq:objective}
\end{equation}
where \( \mathcal{L} \) is a region-aware loss designed to disentangle the modified and unmodified regions, enforcing stability in the unchanged areas while allowing for accurate propagation in the edited regions (details in Sec.~\ref{sec:ra_loss}). To ensure that the final output adheres to real video data distributions, synthetic data is fed exclusively to the content encoder. The I2V generation model, however, uses the original video, preventing the model from inadvertently learning synthetic artifacts.

\subsection{Model Design}
\label{sec:design}
To preserve the unchanged parts of the original video and only propagate the modified regions, we integrate
two additional components to the base I2V model: Selective Content Encoder and Mask Prediction Decoder, as shown in Fig.~\ref{fig:framework-detail}.

\vspace{0.5em}
\noindent\textbf{Selective Content Encoder.} The architecture of our SCE is a replicated version of the initial \( N \) blocks of the main generation model, similar to ControlNet~\cite{zhang2023adding}.
After each encoder block, 
the extracted features are added
to the corresponding features in the I2V model, allowing a smooth and hierarchical flow of content information.
The injection layer is one multilayer perceptron with zero initialization which will also be trained.
Furthermore, for bidirectional information exchange, the features of the I2V model are fused with the SCE's input before the first block.
This lets SCE be aware of the modified regions so that it can selectively encode the information in the unchanged region as intended.

\vspace{0.5em}
\noindent\textbf{Mask Prediction Decoder.} The Mask Prediction Decoder (MPD) is designed to estimate the spatial regions requiring editing, helping the encoder disentangle changes from the unchanged content.
While SCE utilizes the initial \( N \) blocks of the I2V model, MPD mirrors this by using the final block along with one multilayer perceptron (MLP) as the final layer. It takes the latent representation from the penultimate block, which contains rich spatial and temporal information, and processes it through the MLP layer. 
This restores the temporal dimension, matching it to the number of video frames.
The final output is trained to match the instance mask of the video via an MSE loss~\cite{error2010mean} $\mathcal{L}_{\text{MPD}}$.
This guides
the model to focus on the edited regions
and significantly improves the accuracy of the attention maps.

\begin{figure}
  \centering
   \includegraphics[width=.8\linewidth]{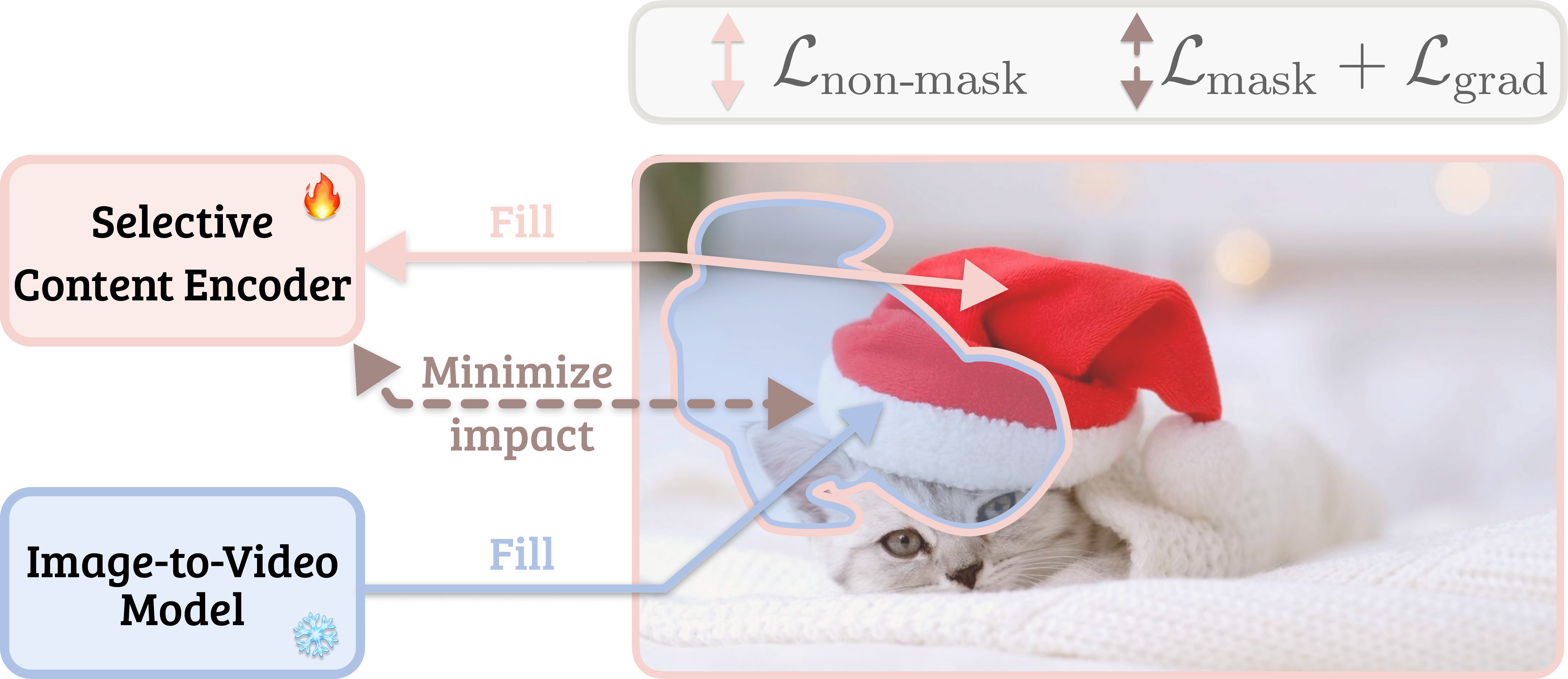}

   \caption{Region-Aware Loss. This loss helps the model to disentangle the edited region from the original content.
   }
   \label{fig:loss}
\end{figure}

\subsection{Region-Aware Loss}
\label{sec:ra_loss}
In our training process, we use instance segmentation data to ensure that both the edited and unedited regions receive appropriate supervision. We design a Region-Aware Loss (RA Loss), shown in Fig.~\ref{fig:loss}, to balance the loss of both regions,
even when the edited areas are proportionally small.

For an input video \( \hat{V} = \{\hat{v}_1, \hat{v}_2, \dots, \hat{v}_T\} \) and instance-level masks \( M = \{m_1, m_2, \dots, m_T\} \), where \( m_t \in \{0, 1\}^{H \times W} \) indicates edited regions in frame \( \hat{v}_t \), we 
apply Gaussian downsampling over the spatial dimensions and repeat over the temporal dimension
to obtain a mask \( \tilde{m}_t \) 
that is aligned to the shape of the
latent representation of the video. 
The loss is separately computed for the mask and non-mask region, giving:
\begin{equation}
\mathcal{L}_{\text{mask}} = \mathbb{E}_{t \sim \mathcal{U}(1, T)} \left[ \mathcal{L}_{\text{d}}(\tilde{m}_t \cdot v_t^{\text{out}}, \tilde{m}_t \cdot v_t) \right] \text{and}
\end{equation}
\[
\mathcal{L}_{\text{non-mask}} = \mathbb{E}_{t \sim \mathcal{U}(1, T)} \left[\mathcal{L}_{\text{d}}((1 - \tilde{m}_t) \cdot v_t^{\text{out}}, (1 - \tilde{m}_t) \cdot v_t) \right],
\]
where \( \mathcal{L}_{\text{d}} \) denotes the diffusion MSE loss that measures the pixel-wise error between the generated frame \( v_t^{\text{out}} \) and ground truth \( v_t \).

To further reduce the SCE’s influence on the masked regions, we add a gradient loss \( \mathcal{L}_{\text{grad}} \) that minimizes the effect of the masked area in the encoder’s input. Instead of computing second-order gradients, we approximate using a finite difference:
\begin{equation}
\Delta f = \frac{f(\mathcal{E}(\hat{V} + \delta)) - f(\mathcal{E}(\hat{V}))}{\delta}
\end{equation}
where \( f(\mathcal{E}(\hat{V})) \) represents the encoder’s feature, and \( \delta \) is a small perturbation. The gradient loss is defined as:
\begin{equation}
\mathcal{L}_{\text{grad}} = \mathbb{E}_{t \sim \mathcal{U}(1, T)} \left[ \tilde{m}_t \cdot \left\| \Delta f \right\|_2 \right].
\end{equation}
The RA Loss $\mathcal{L}$ is a weighted sum of all three terms to ensure sufficient supervision on both masked and unmasked areas:
\begin{equation}
\mathcal{L} = \mathcal{L}_{\text{non-mask}} + \lambda \cdot \mathcal{L}_{\text{mask}} + \beta \cdot \mathcal{L}_{\text{grad}} + \gamma \cdot \mathcal{L}_{\text{MPD}}
\end{equation}
 
\subsection{Synthetic Data Generation}
\label{sec:data_gen}
Creating a large-scale paired video dataset can be costly and challenging especially for video propagation, as it is difficult to encompass all video tasks. To address this, we propose to use synthetic data derived from video instance segmentation datasets.
In our training, we use Youtube-VOS~\cite{xu2018youtube}, SAM-V2~\cite{ravi2024sam2}, and an internal dataset. However, this data generation pipeline can be applied to any available video instance segmentation dataset.
Specifically, we adopt a mix of augmentation techniques to the segmentation data, tailored to various propagation sub-tasks: (1) \textit{Copy-and-Paste}: Objects from one video are randomly segmented and pasted into another, simulating object insertion; (2) \textit{Mask-and-Fill}: The masked region undergoes inpainting, creating realistic edits within selected regions; (3) \textit{Color Fill}: The masked area is filled with specific colors, representing basic object tracking scenarios. For (3), $V$ will be sent to $\mathcal{E}$ and $\hat{v}_1$ will be sent to $\mathcal{G}$ in Eq.~\ref{eq:objective}.
Each synthetic data type aligns with a distinct task, enabling our model to generalize across diverse applications.
Task embeddings corresponding to these 
augmentation methods are injected into the model, guiding the model to adapt based on the augmentation type. Note that despite the variety of data creation methods and tasks, the core function of SCE remains consistent: 
encode the unedited information while the I2V model maintains the generative capabilities to propagate the edited regions.
More details about each augmentation technique are provided in the Supplementary Material.
\section{Experiments}
\label{sec:exp}

\begin{figure*}
  \centering
   \includegraphics[width=0.96\linewidth]{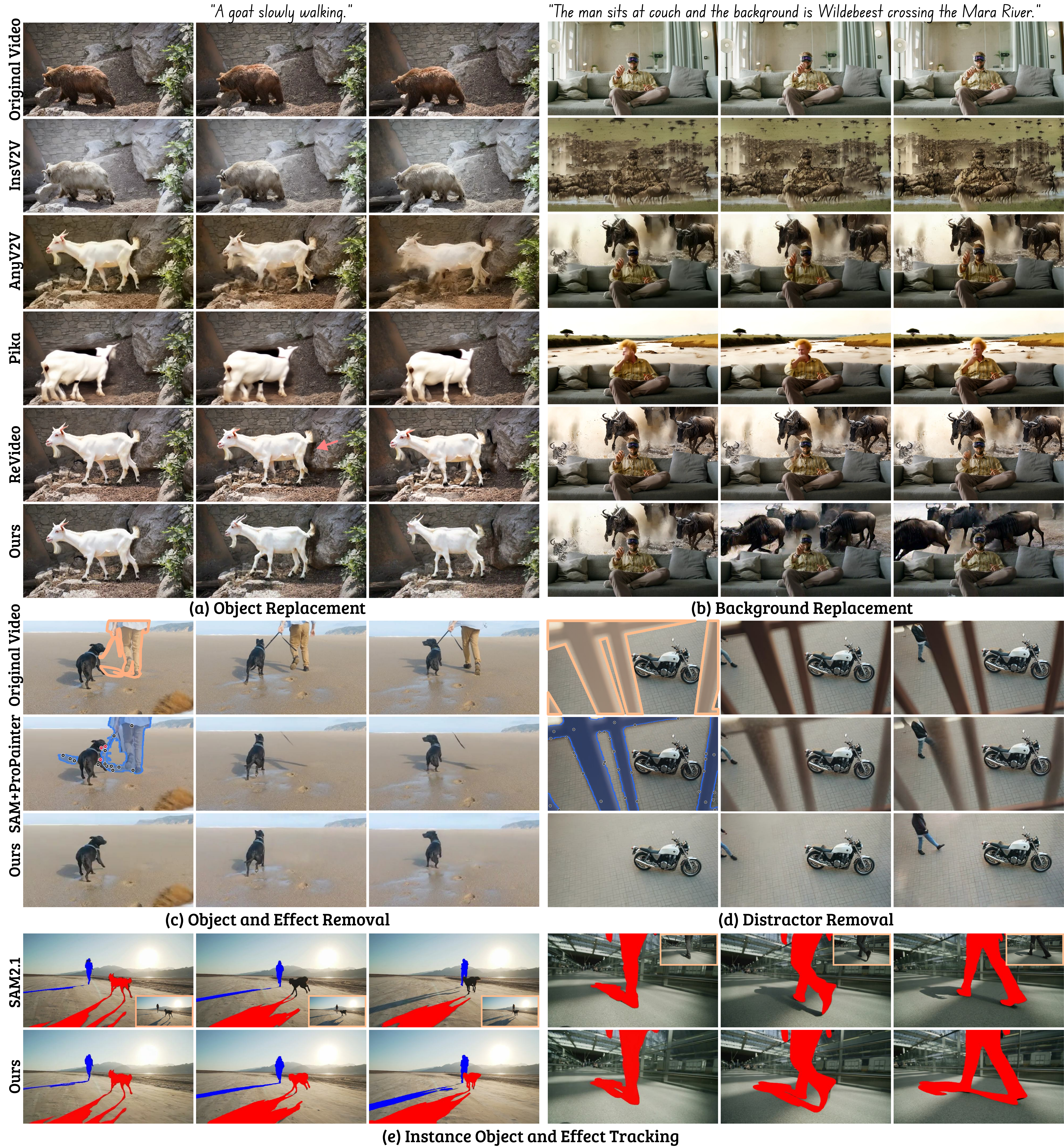}

   \caption{Visual Comparison in Multiple Video Tasks. GenProp demonstrates versatile editing capabilities, (a) allowing seamless modification of objects into those with vastly different shapes with independent motion and (b) enabling background edits. For object removal, GenProp excels at (c) effectively removing object effects together with the object and (d) realistically reconstructing large occluded areas. It is further able to perform instance tracking of objects and their effects when solid color fills are given as the first frame (see (e)).}
   \vspace{-2mm}
   \label{fig:qual}
\end{figure*}

\begin{table*}[t]
  \centering
  \scalebox{0.83}{
  \begin{tabular}{l | c c c | c c | c c c | c c}
    \toprule
    \multirow{3}{*}{Method} & \multicolumn{5}{c|}{Classic Test Set} & \multicolumn{5}{c}{Challenging Test Set} \\
    & \multirow{2}{*}{$\text{PSNR}_m$ $\uparrow$} & \multirow{2}{*}{CLIP-T $\uparrow$} & \multirow{2}{*}{CLIP-I $\uparrow$} & \multicolumn{2}{c|}{GenProp preference \%} & \multirow{2}{*}{$\text{PSNR}_m$ $\uparrow$} & \multirow{2}{*}{CLIP-T $\uparrow$} & \multirow{2}{*}{CLIP-I $\uparrow$} & \multicolumn{2}{c}{GenProp Preference \%} \\
    &&&& Alignment & Quality &&&& Alignment & Quality \\
    \midrule
    InsV2V~\cite{cheng2023consistent} & 28.999 & 0.3049 & 0.9737 & 60.00 & 60.00 & 28.842 & 0.2906 & 0.9718 & 81.82 & 75.00 \\
    AnyV2V~\cite{ku2024anyv2v} & 32.090 & 0.3050 & 0.9676 & 95.56 & 86.67 & 28.338 & 0.3302 & 0.9576 & 97.78 & 95.56 \\
    Pika~\cite{pika} & 32.568 & 0.3226 & \textbf{0.9923} & 62.22 & 55.56 & 31.329 & 0.3023 & 0.9886 & 88.89 & 86.67 \\
    ReVideo~\cite{mou2024revideo} & 31.765 & 0.3196 & 0.9777 & 75.56 & 71.11 & 29.920 & 0.3226 & 0.9798 & 84.44 & 82.22 \\
    \textbf{GenProp (Ours)} & \textbf{33.837} & \textbf{0.3229} & 0.9825 & - & - & \textbf{32.163} & \textbf{0.3336} & \textbf{0.9904} & - & - \\
    \bottomrule
  \end{tabular}
  }
  \caption{Video editing benchmark compared to existing models. $\text{PSNR}_m$ measures the consistency outside the edited region and Text Alignment and Consistency metrics measure the edit quality. User study shows the percentage of users who preferred Ours over the compared method on alignment (left) and quality (right). GenProp significantly outperforms the other methods on the Challenging Set.}
  \label{tab:edit-complex}
\end{table*}

\begin{table}[t]
  \centering
  \scalebox{0.85}{
  \begin{tabular}{l | c | c c }
    \toprule
    \multirow{2}{*}{Method} & \multirow{2}{*}{CLIP-I $\uparrow$} & \multicolumn{2}{c}{GenProp Preference \%} \\
    && Alignment & Quality \\
    \midrule
    SAM + Propainter & 0.9809 & 82.22 & 75.56 \\
    ReVideo~\cite{mou2024revideo} & 0.9728 & 86.36 & 77.27 \\
    \textbf{GenProp (Ours)} & \textbf{0.9879} & - & - \\
    \bottomrule
  \end{tabular}
  }
  \caption{Object removal comparison to other methods. GenProp outperforms baselines on consistency, alignment, and quality.}
  \label{tab:remove}
\end{table}

\subsection{Implementation Details}
As GenProp is a general framework, we experiment with both a DiT architecture similar to Sora~\cite{sora} and a U-Net architecture based on Stable Video Diffusion (SVD)~\cite{blattmann2023stable} as the base video generation model. For the former, it is trained
for I2V generation on 32, 64, and 128 frames at 12 and 24 FPS, with a base resolution of 360p.
SCE (24 blocks) and MPD are trained while the I2V model is frozen.
The results can be upscaled to 720p using a super-resolution model.
The learning rate is set to 5e-5 with a cosine-decay scheduler and a linear warmup. An exponential moving average is applied for training stability. A gradient norm threshold of 0.001 prevents training instability. Classifier-free guidance (CFG) value is set to 20, and the data augmentation ratio is set to 0.5/0.375/0.125 for copy-and-paste/mask-and-fill/color fill. In the RA loss, $\lambda$ is 2.0, $\beta$ is 1.0, and $\gamma$ is 1.0.
All experiments were conducted on 32/64 NVIDIA A100 GPUs for different architectures. 
We find that the DiT backbone has a better video generation quality. Our main results are from this DiT variant while the ablation studies are conducted with the SVD-based architecture. Please refer to the Supplementary Material for the results based on SVD.

\subsection{Comparisons}
As generative video propagation is a new problem, we compare the SotA methods in each of the three sub-tasks of GenProp. Note that our model is able to handle these tasks within the same model and further cover additional tasks such as outpainting as well as combinations of these sub-tasks as shown in the bottom row of Fig.~\ref{fig:teaser}. We provide extensive results in the Supplementary Material.

\vspace{-4mm}
\paragraph{Diffusion-based Video Editing}
In Fig.~\ref{fig:qual} (a) and (b), we compare GenProp with other diffusion-based video editing methods, including text-guided and image-guided approaches. InsV2V~\cite{cheng2023consistent} relies on instruction text for controlling generation. However, due to its limited training data, it struggles with significant shape changes and does not support object insertion. Pika~\cite{pika} also uses text prompts to edit within a box region, but it performs poorly when the object’s shape changes substantially and cannot handle background edits or object insertion.
AnyV2V~\cite{ku2024anyv2v} is a training-free method that uses the first frame to guide editing. While it handles appearance changes, it fails when there are large shape or background modifications, often resulting in degradation or ghosting effects. Like InsV2V and Pika, it also cannot insert objects. We use ReVideo~\cite{mou2024revideo} to manage large shape changes by first removing an object and then re-inserting it, but this two-stage process has drawbacks. The box-based region can cause blurry boundaries, and object motion is affected by the original point tracking, leading to accumulated errors. Additionally, the box region limits its ability to edit complex backgrounds effectively.

\vspace{-4mm}
\paragraph{Video Object Removal}
For object removal, we compare GenProp with 
a traditional inpainting pipeline, where we cascade two SotA models to achieve a propagation-like inpainting, since traditional methods require a dense mask annotation for all frames: SAM-V2~\cite{ravi2024sam2} for mask tracking, then Propainter~\cite{zhou2023propainter} for inpainting the regions in the estimated masks.
As shown in Fig.~\ref{fig:qual} (c) and (d), GenProp has several advantages: 
(1) no need for a dense mask annotation as input;
(2) removal of object effects like reflections and shadows;
(3) removal of large objects and natural filling within large areas.

\vspace{-4mm}
\paragraph{Video Object Tracking}
We compare GenProp with SAM-V2~\cite{ravi2024sam2} on tracking performance in Fig.~\ref{fig:qual} (e). Since SAM-V2 is trained on the large-scale SA-V dataset, it is expected that SAM-V2 often produces more precise tracking masks than GenProp. Additionally, GenProp is slower than real-time tracking methods like SAM-V2. However, it has notable advantages. Due to its video generation pretraining, GenProp has a strong understanding of physical rules. As shown in Fig.~\ref{fig:qual}, unlike SAM-V2, which struggles with object effects like reflections and shadows due to limited and biased training data, GenProp can consistently track these effects. This highlights the potential of approaching classic vision tasks with generation-based models.

\begin{table}[t]
  \centering
  \scalebox{0.85}{
  \begin{tabular}{l | c |c }
    \toprule
    Method & CLIP-T $\uparrow$ & CLIP-I $\uparrow$\\
    \midrule
    w/o MPD & 0.3252 & 0.9834 \\
    w/o RA Loss & 0.3261 & 0.9825 \\
    \textbf{GenProp (Ours)} & \textbf{0.3316} & \textbf{0.9872} \\
    \bottomrule
  \end{tabular}
  }
  \caption{Ablation study. Both MPD and RA loss can improve the success rate of editing and the quality of the output video.}
  \label{tab:ablation}
\end{table}

\vspace{-4mm}
\paragraph{Quantitative Results}
We conduct a quantitative evaluation on several test sets. 
For video editing (reported in Tab.~\ref{tab:edit-complex}), we evaluate on two types of test sets: (1) Classic Test Set, which is TGVE~\cite{wu2023cvpr}'s DAVIS~\cite{davis2017pont} part and its ``Object Change Caption'' as the text prompt, focusing on object replacement and appearance editing; (2) Challenging Test Set, which is 30 manually collected videos from Pexels~\cite{pexels} and Adobe Stock~\cite{stock} including large object replacement, object insertion and background replacement. For (2), the first frame is edited using a commercial photo editing tool.
For Pika \cite{pika}, we use the online boxing tool, running it three times for each result. For ReVideo \cite{mou2024revideo}, we select a box region, then to track appearance changes, we use its code to extract the original object's motion points. 
For edits with significant shape changes, we first remove the original object
and then insert the new object, assigning a future trajectory.
For assessing the consistency in the unchanged regions, we measure the PSNR outside the edit mask, denoted as $\text{PSNR}_m$.
For cases with large shape changes, we apply a rough mask over the original and edited regions, only calculating the PSNR on areas outside these masks. For text alignment, we compute the cosine similarity between the CLIP~\cite{radford2021learning} embeddings of the edited frame and the text prompt (CLIP-T) \cite{wu2023cvpr,mou2024revideo,ouyang2024i2vedit}. For result quality, we calculate the distance between CLIP~\cite{radford2021learning} features across frames (CLIP-I) \cite{wu2023cvpr,mou2024revideo,ouyang2024i2vedit}. As shown in Tab.~\ref{tab:edit-complex}, GenProp outperforms the other methods on most metrics, especially on the Challenging Test Set. Pika exhibits better consistency on the Classic Test Set, as its bounding box performs reasonably well when object shapes remain relatively unchanged. ReVideo degrades on multiple objects.

For object removal, we collect 15 videos with complex scenes, including object effects and occlusions, as existing test sets lack coverage of these cases. For SAM, we click on the object and side effects to ensure complete coverage. As shown in Tab.~\ref{tab:remove}, GenProp achieves the highest consistency, while ReVideo may produce bounding box artifacts, and ProPainter struggles with object effects.

As quality metrics often do not correctly capture the realism of the generated results,
we use Amazon MTurk~\cite{turk2012amazon} to conduct a user study with a total of 121 participants. Each participant views several videos generated by GenProp and a random baseline, along with the original video and the text prompt. They are asked two questions: 1) Which video aligns better with the instructions? 2) Which video is visually better? Participants then select one video for each question.
In Tables~\ref{tab:edit-complex} and ~\ref{tab:remove}, we show the percentage of time users prefer Ours over the competing baselines (alignment/quality).
GenProp outperforms all baselines by a large margin, especially on the Challenging Test Set.

\begin{figure}
  \centering
   \includegraphics[width=1\linewidth]{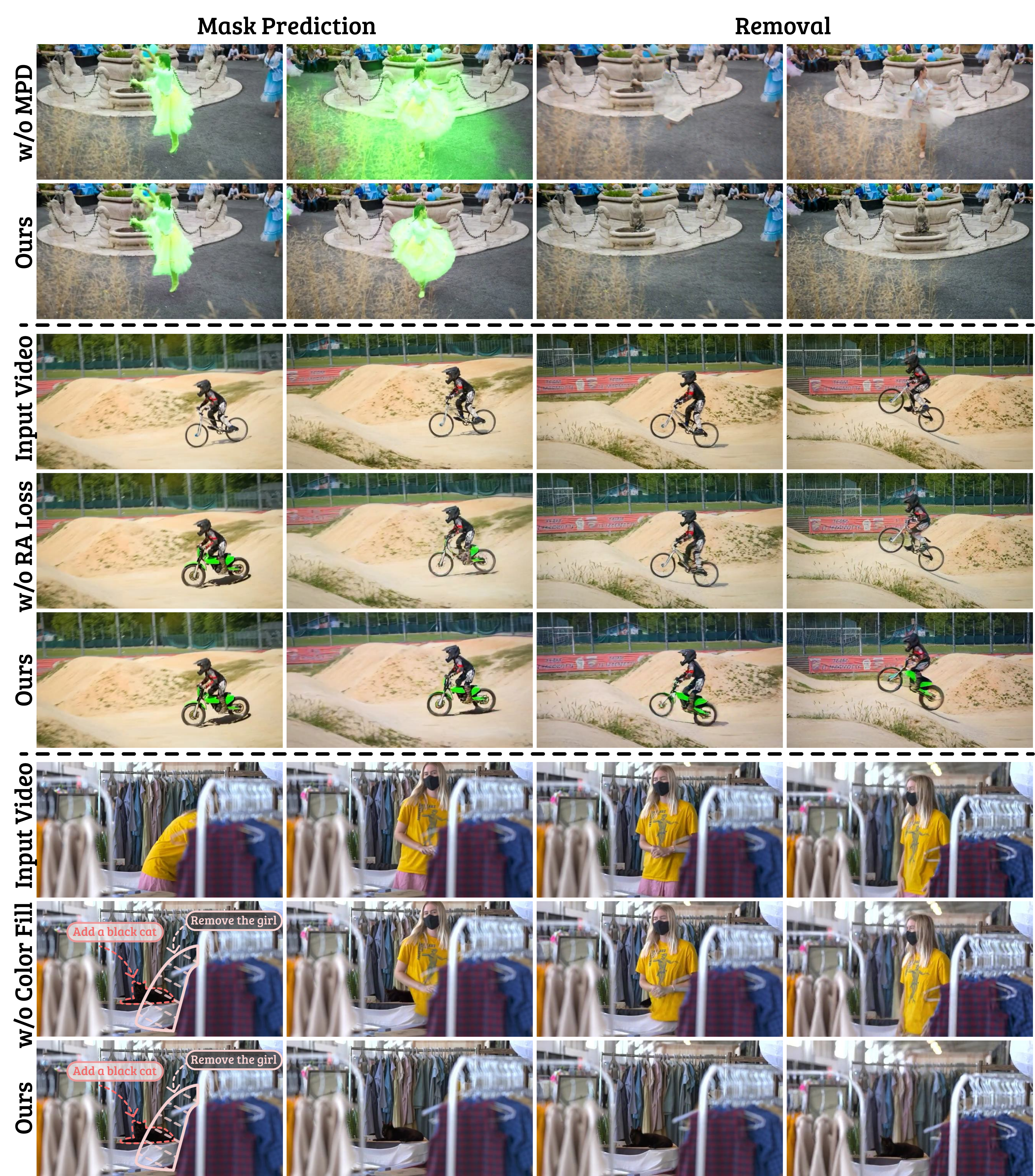}
   \caption{Visual comparison of model variants, showing the effect of MPD (top), RA loss (middle) and Color Fill (bottom).}
   \vspace{-1mm}
   \label{fig:qual-ablation}
\end{figure}

\subsection{Ablation Study}

\paragraph{Mask Prediction Decoder}
In Tab.~\ref{tab:ablation}, we evaluate the effect of MPD on the Challenging Test Set, showing that it can improve both Text Alignment and Consistency. As shown in Fig.~\ref{fig:qual-ablation} rows 1 and 2, without MPD, the output mask is often highly degraded, leading to worse removal quality. 
Without explicit supervision with MPD, the model may be confused which part to propagate and which part to preserve in the original video, causing partially removed objects to reappear in the following frames. MPD helps the disentanglement and both the removal results and predicted masks become more accurate with MPD, allowing for full object removal even with heavy occlusion.

\vspace{-4mm}
\paragraph{Region-Aware Loss}
In Tab.~\ref{tab:ablation}, we further test the effectiveness of the proposed RA Loss on the Challenging Test Set. A core challenge in GenProp is that SCE can mistakenly select all regions from the original video including the edited areas, weakening the I2V generation ability due to the reconstruction loss. As shown in Fig.~\ref{fig:qual-ablation} rows 3-5, without RA Loss, the original object tends to gradually reappear, 
hindering the propagation of the first-frame edit (the green motor).
With RA Loss, the edited areas are able to be propagated in a stable and consistent way.

\vspace{-4mm}
\paragraph{Color Fill Augmentation}
Color Fill augmentation is another crucial factor for addressing the propagation failure.
While copy-and-paste and mask-and-fill augmentations allow the model to implicitly learn object modifications, replacements, and deletions, color filling explicitly trains it for tracking, guiding the model to maintain modifications made in the first frame throughout the sequence, with the prompt ``track colored regions''. As shown in Fig.~\ref{fig:qual-ablation} rows 6-8, changing the girl into a small cat is challenging due to the significant shape difference. However, with color fill augmentation, GenProp successfully propagates this large modification throughout the sequence.
\section{Conclusion}
\label{sec:con}
In this paper, we design a novel \textit{generative video propagation} framework, GenProp, that harnesses the inherent video generation power of I2V models to achieve various downstream applications
including removal, insertion and tracking.
We demonstrate its potential by showing that it is able to expand the range of achievable edits (e.g., remove or track objects together with their associated effects) and generate highly realistic videos,
without relying on traditional intermediate representations like optical flow or depth maps. By integrating a selective content encoder and leveraging an I2V 
generation model, GenProp consistently preserves unchanged content while dynamically propagating the changes. 
Synthetic data and the region-aware loss further enhance its ability to disentangle and refine edits across frames. Experimental results demonstrate its effectiveness, establishing it as a robust, flexible solution that surpasses prior methods in scope and precision. In the future, we plan to extend the model to take in more than one key frame edits and uncover additional video tasks that can be supported.

\clearpage

{
    \small
    \bibliographystyle{ieeenat_fullname}
    \bibliography{main}
}

\clearpage
\setcounter{page}{1}
\onecolumn
\setcounter{section}{0} 
\renewcommand{\thesection}{S\arabic{section}}
\section*{\centering Supplementary Material}

\section{Synthetic Data Generation}
Our model (GenProp) is trained on synthetic data derived from video instance segmentation datasets. The synthetic data pairs are generated using a combination of methods: (1) Copy-and-Paste for object removal, (2) Mask-and-Fill for editing, and (3) Color-Fill techniques for tracking. These methods ensure diverse training scenarios while maintaining control over the generated content.

\subsection{Copy-and-Paste}
To generate synthetic training data, we employ a copy-and-paste strategy in the dataloader. For each iteration, two videos $V_1 = (v_{1,1}, \dots, v_{1,n})$ and $V_2$ are sampled. We check whether $V_2$ contains an instance mask in the first frame, as our model modifies the video based on the first frame. If neither video has an instance mask in the first frame, the sample is skipped.

Otherwise, the augmented video $V_{\text{aug}}$ is created as:  
\begin{equation}
    V_{\text{aug}} = (1 - \text{M}_2) \odot V_1 + \text{M}_2 \odot V_2,
\end{equation}
where $\text{M}_2$ represents the instance mask of $V_2$, and $\odot$ denotes element-wise multiplication. This operation pastes the object from $V_2$ onto $V_1$.

As illustrated in Fig.~\ref{fig:data} (a), rows 1--6, this approach is simple and efficient, enabling rapid generation of large-scale synthetic data. However, it does not explicitly address harmonization between the pasted object and the target video. The size, position, and motion trajectory of the pasted object vary.

\subsection{Mask-and-Fill}
For the Mask-and-Fill strategy, a single video $V = (v_1, \dots, v_n)$ is sampled at each iteration. Similar to the copy-and-paste strategy, we ensure that the first frame contains an instance mask. If no mask is present in the first frame, the sample is skipped. To fill the instance mask, we employ two approaches:

\paragraph{Surrounding Background Mean Fill}  
This method fills masked regions using the mean pixel value of a rectangular area surrounding the mask, as shown in Fig.~\ref{fig:data}~(b), rows 1--2. For each frame, the bounding box of the mask is identified and expanded by a margin of $5$ pixels. The mean pixel value of the unmasked region within this area is then computed and used to replace the masked region. This approach is simple and efficient, providing a quick solution for local content replacement or insertion.

\paragraph{OpenCV-Based Inpainting}  
As shown in Fig.~\ref{fig:data}~(b), rows 3--4, this method utilizes OpenCV's \texttt{cv2.inpaint()} function with the \texttt{INPAINT\_TELEA} algorithm. The algorithm~\cite{telea2004image} reconstructs the masked regions by interpolating from the surrounding pixels.

Both methods are lightweight and designed for real-time data generation, allowing synthetic data to be processed online during training. Surrounding Background Mean Fill prioritizes simplicity and speed, while OpenCV-Based Inpainting offers more sophisticated results at a slightly higher computational cost. The ratio between the two methods is approximately 2:1.

\subsection{Color-Fill}
In this method, the segmentation masks are used to directly fill occluded regions with a predefined color. The default color is red (R=1.0, G=0.0, B=0.0), but a random color is sampled from a predefined palette, including green, blue, yellow, purple, and cyan. Specifically, given a binary segmentation mask, regions marked with ``1'' are replaced with the randomly selected color, while regions marked with ``0'' are preserved from the original frame. In 30\% of the cases, a second color is randomly sampled for another instance, promoting the model’s ability to track multiple instances.

\begin{figure*}
  \centering
   \includegraphics[width=.96\linewidth]{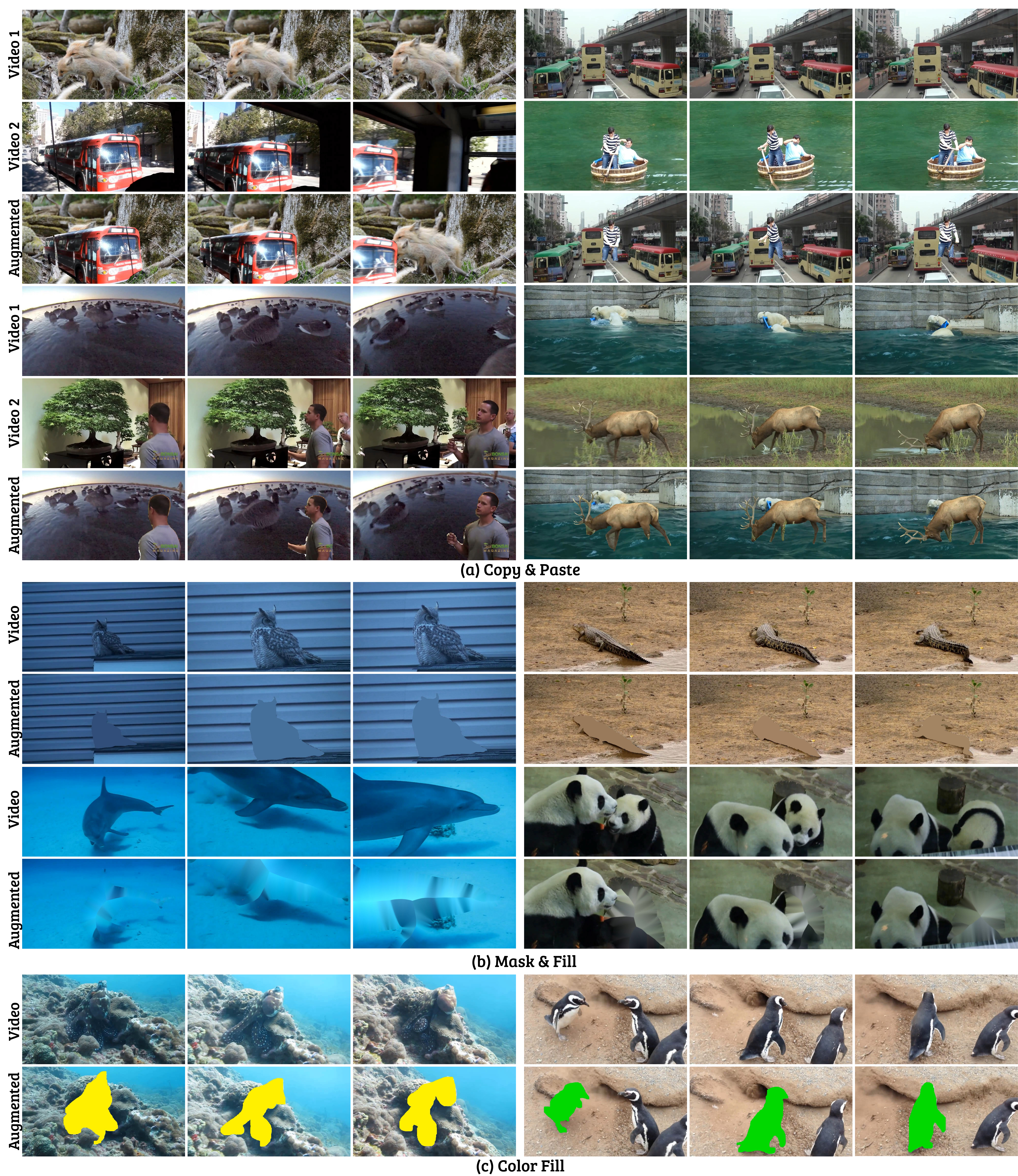}
   \caption{Synthetic Data Generation. We use different ways to generate our training data by simulating a task: (a) Copy-and-Paste for object removal; (b) Mask-and-Fill for editing and insertion; (c) Color Fill for both tracking and editing enhancement.}
   \vspace{-1mm}
   \label{fig:data}
\end{figure*}

The procedure is straightforward yet effective, as it introduces strong visual cues that highlight the areas where propagation tasks occur. As illustrated in Fig.~\ref{fig:data} (c), this method is particularly useful for training tasks that require tracking or editing specific regions, as the distinct colors ensure clear differentiation of object instances across frames.

\clearpage

\section{Controls for Generation}

\begin{figure*}
  \centering
   \includegraphics[width=.96\linewidth]{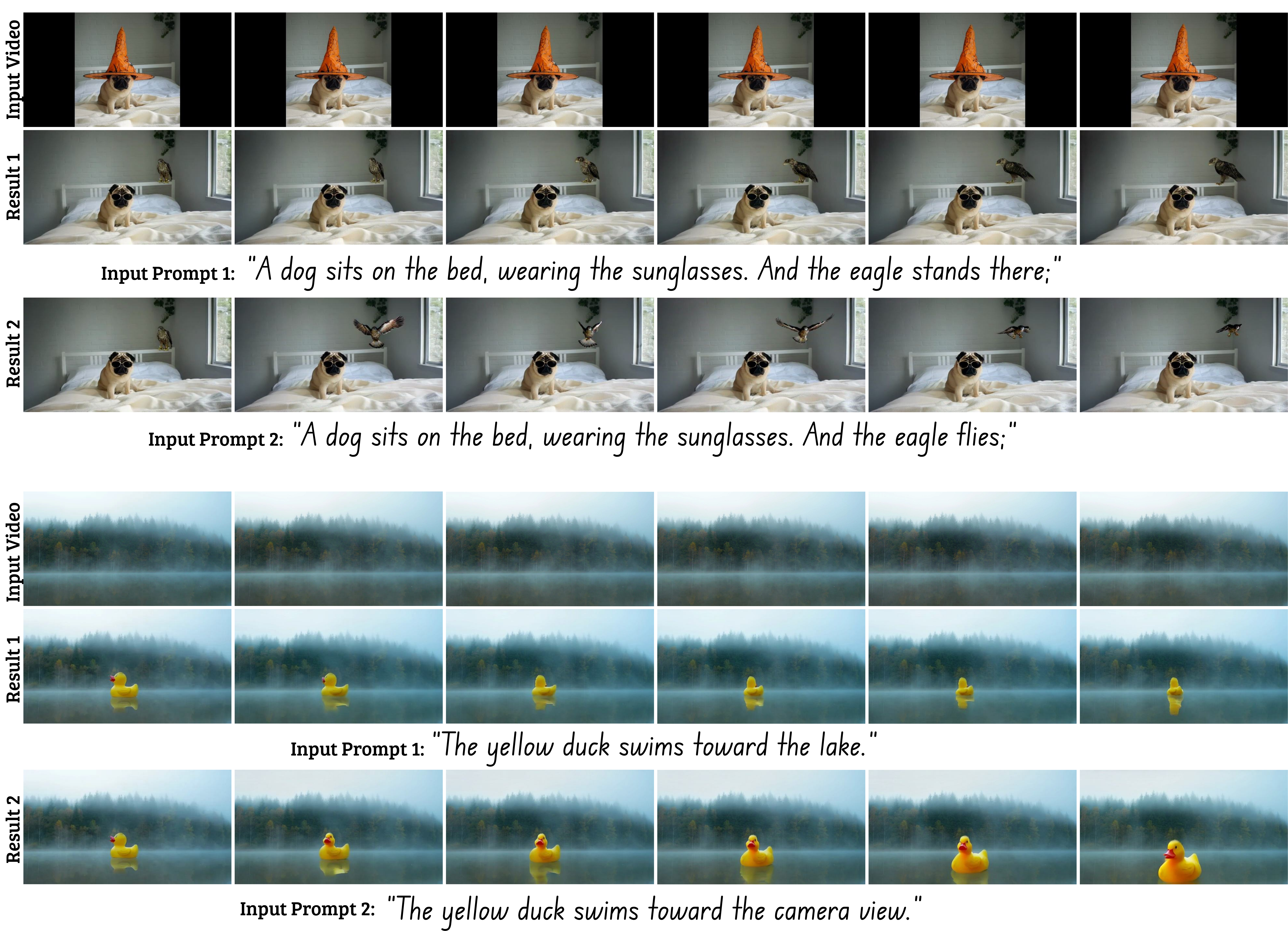}
   \caption{Text Control Analysis. Text prompts can be used to control the result in a desired way.}
   \vspace{-1mm}
   \label{fig:text}
\end{figure*}

\subsection{Text Control Analysis}
In GenProp, the text prompt also plays a role in guiding the model to generate content that aligns with the desired outcome. The interaction between the edited first frame and the input video, combined with the provided text prompt, results in different outputs, demonstrating the potential influence of text control on video propagation.

In Fig.~\ref{fig:text} rows 1-3, we illustrate a scenario involving multiple edits, including object removal and editing. In this example, an eagle is inserted into the video, and the text prompt is used to control the eagle's behavior—whether it ``stands'' or ``flies''. The text prompt directs how the eagle is depicted and how it moves within the video.

In Fig.~\ref{fig:text} rows 4-6, we show a video of a lake surface with mist, where a small yellow duck is inserted in the first frame. By varying the text prompt, the direction in which the duck swims can be controlled. Different text prompts guide the duck’s movement, demonstrating the model's ability to follow text cues for spatial and motion control, adding an extra layer of flexibility for dynamic video editing tasks.

These examples underscore the capacity of GenProp to integrate textual instructions effectively, allowing for nuanced and adaptable control over the generated video content, making it a powerful tool for both creative video editing and dynamic scene manipulation.

\subsection{Injection Weight Analysis}
\begin{figure*}
  \centering
   \includegraphics[width=.96\linewidth]{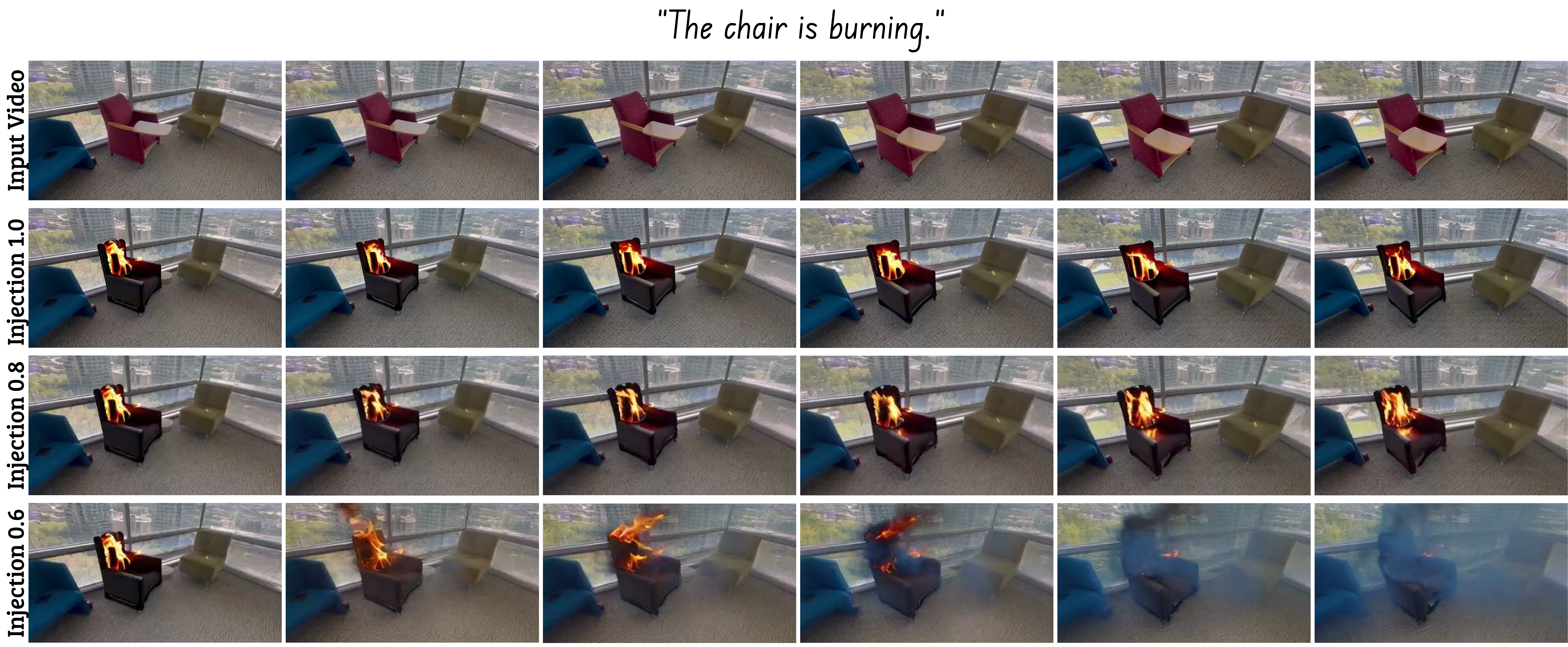}
   \caption{Injection Weight Analysis. The injection weight serves as a way to control the trade-off between reconstruction ability and generation ability. With a lower injection weight, edits with significant changes can appear in the original video as shown in the last row.}
   \vspace{-1mm}
   \label{fig:inject}
\end{figure*}

As shown in Fig.~\ref{fig:inject}, the injection layer connects the output of the Select Content Encoder (SCE) to the Image-to-Video Model, enabling the selective propagation of content between the original video and the generated edits. To control the balance between preserving the original video and generating the edited content, we introduce an injection weight parameter, ranging from [0.0, 1.0], multiplied by the injection layer, which can be adjusted during the inference phase. This injection weight serves as a trade-off, allowing for more control over how much of the original video is reconstructed versus how much of the newly generated content is introduced.

For instance, as shown in Fig.~\ref{fig:inject}, we use a video of a sofa and edit the first frame to make it appear as if it is burning. When the injection weight is set to 1.0, the reconstruction of the original video is highly accurate, but the flame effects are relatively small. As the injection weight is decreased to 0.8, the flames become more pronounced while still maintaining a strong reconstruction of the original content. At an injection weight of 0.6, the reconstruction of the ground and windows is somewhat weakened, but the generated smoke from the flames can spread over a much larger area, showcasing how the injection weight directly influences the extent to which the model prioritizes either reconstruction or generation of new content.

\subsection{Black Region in Input Video}
In the standard GenProp setting, the Selective Content Encoder (SCE) takes the original video as input. The SCE's task is to distinguish between modified and unmodified content. Adding appropriate masks to the input video can help the SCE focus on this task and improve the model's overall performance.
We also found that using moving masks in the input video can guide the motion of the modified content. This provides a certain level of control over the motion of the edited regions.

\begin{figure*}
  \centering
   \includegraphics[width=.99\linewidth]{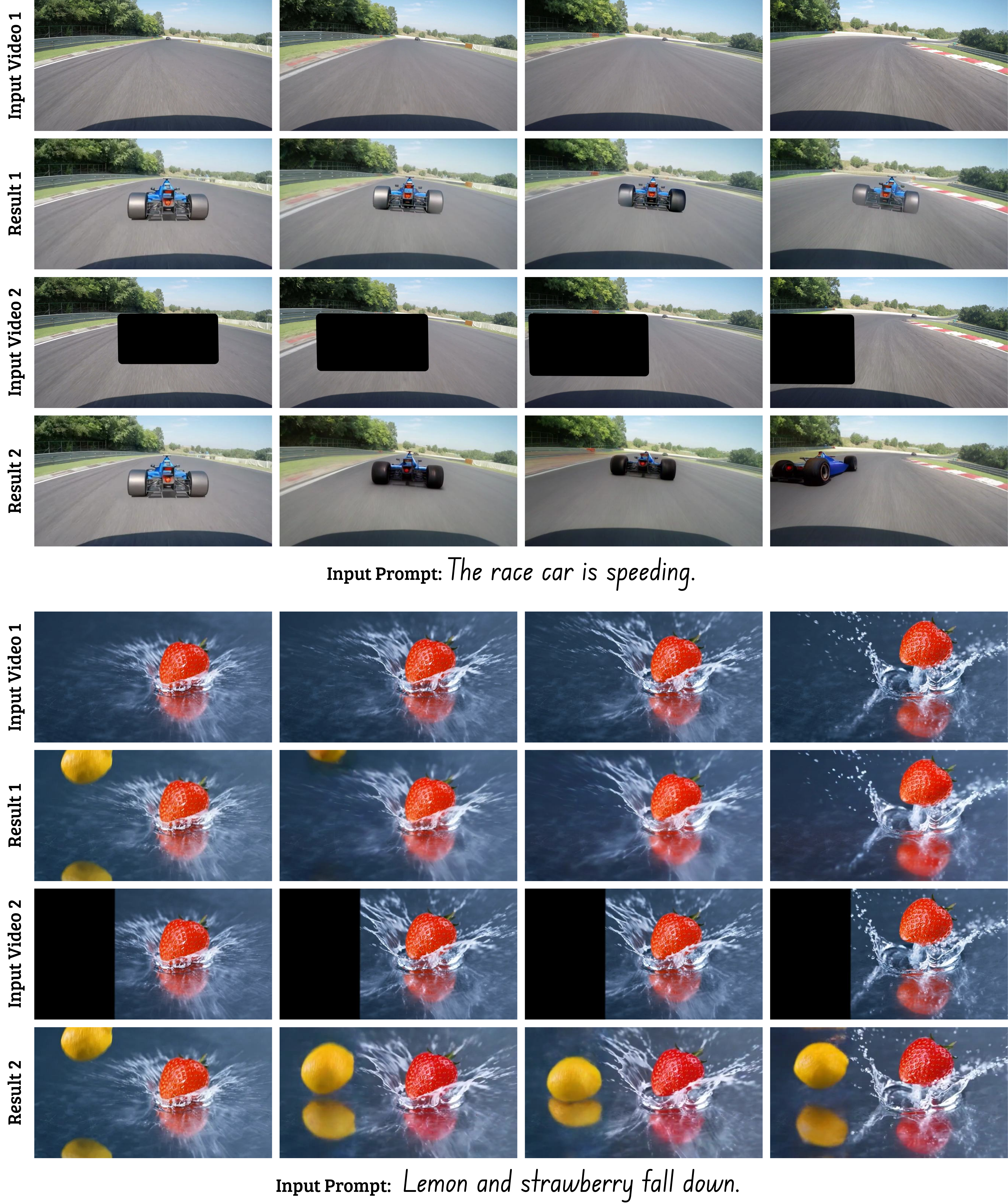}
   \caption{Motion Control with Black Regions. Adding back regions to the input video can help to control the motion of the element we want to edit in the video. For example, we can simulate overtaking of a racecar (top) or make the lemon fall to the left of the strawberry (bottom).}
   \vspace{-1mm}
   \label{fig:blackregion}
\end{figure*}

Fig.~\ref{fig:blackregion} demonstrates that adding a black region to the input video can help control the motion of the element we want to edit. Specifically, in the first case, we can use the moving black blocks in the input video to simulate the effect of a car being overtaken. In the second case, the black region helps the model to use text to control the motion of the lemon.

\clearpage
\section{User Study Details}
\begin{figure*}
  \centering
   \includegraphics[width=.9\linewidth]{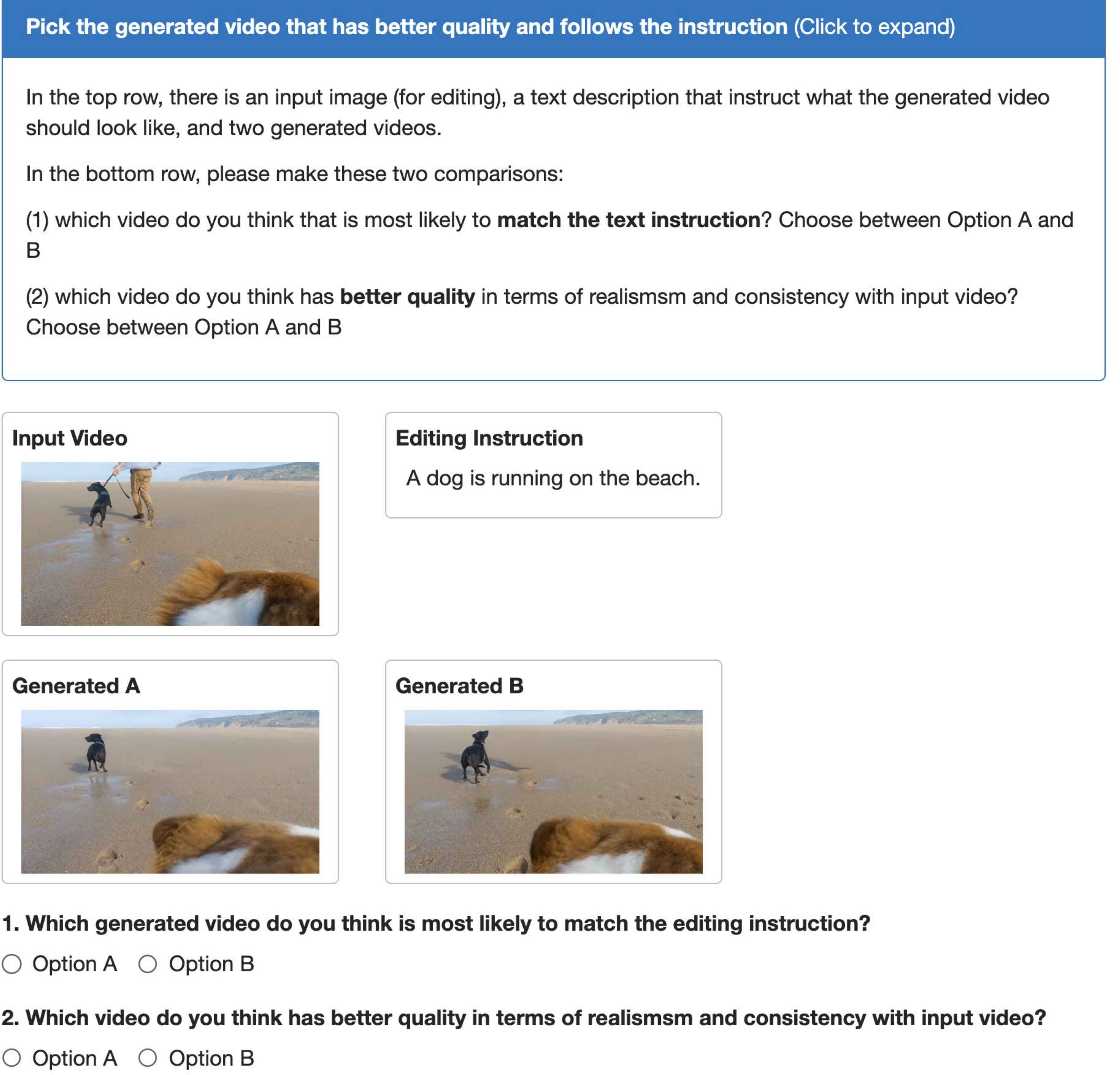}
   \caption{User Study Interface. Screenshot of a user study screen where two questions are asked to the user for assessing (1) alignment to the text and (2) overall video quality.}
   \vspace{-1mm}
   \label{fig:user-study}
\end{figure*}

Fig.~\ref{fig:user-study} shows the interface used in our user study. In this study, users are presented with an input video, a corresponding text prompt, and the results generated by both our GenProp model and a random baseline (with users unaware of which result corresponds to which model). The users are asked to evaluate the outputs based on two criteria: ``alignment to the editing goal'' and ``output video quality''. Specific questions related to these criteria are detailed in the figure.
At the end of the study, participants' responses are collected in a CSV format. To ensure the reliability of the results, we perform a systematic filtering of user responses, excluding those from participants who exhibited unreasonable response times (less than 1 second), ensuring that the data reflects thoughtful and accurate assessments.
This user study setup allows us to compare the performance of GenProp against a baseline and gain insights into the effectiveness of our model in real-world editing tasks.

\clearpage

\begin{figure}[tp]
  \centering
   \includegraphics[width=.6\linewidth]{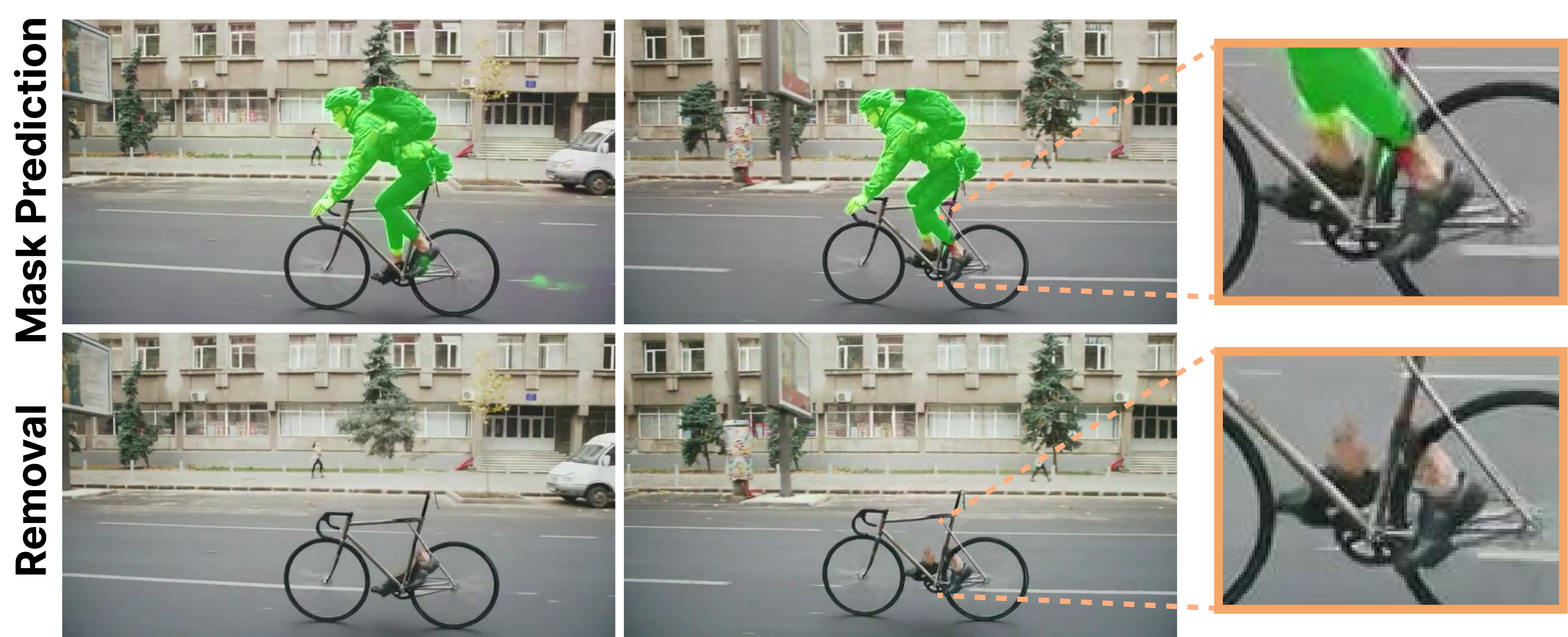}

   \caption{Observation. When the mask prediction fails, the editing may fail in a similar manner.}
   \label{fig:track&edit}
\end{figure}

\begin{figure}
  \centering
   \includegraphics[width=.6\linewidth]{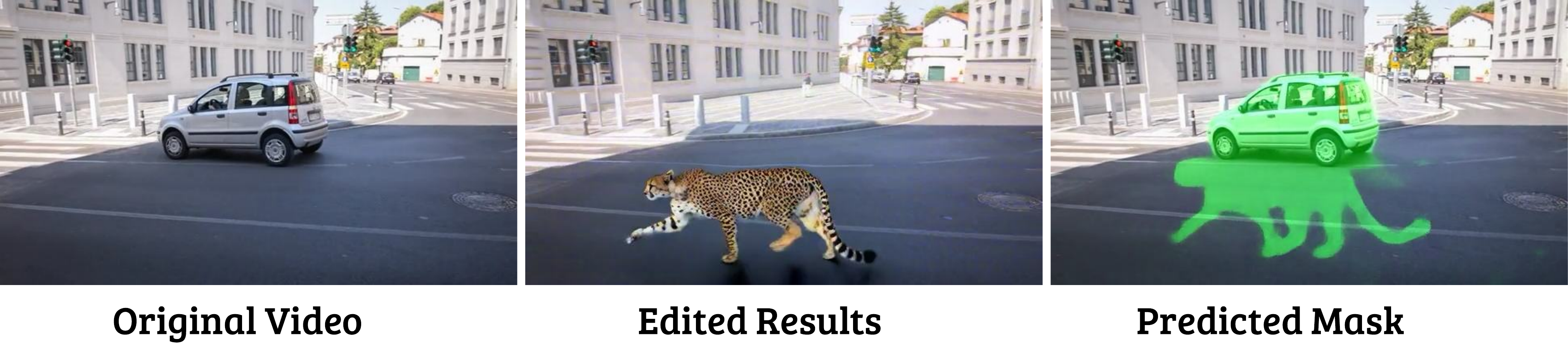}

   \caption{Mask Visualization. Mask prediction decoder can estimate the edited region, even when its shape extends beyond the original object.}
   \label{fig:edit-mask}
\end{figure}

\section{Mask Prediction Analysis}
For the Mask Prediction Decoder (MPD), we make additional observations. As shown in Fig.~\ref{fig:track&edit}, the editing outcomes and the mask prediction results often succeed or fail in the same way. This correlation highlights the importance of accurate mask predictions for generating high-quality edits.
As further illustrated in Fig.~\ref{fig:edit-mask}, MPD is not only capable of predicting the object that is removed from the original video (which it is trained to) but can also estimate its effect (shadow) and the future appearance areas of inserted objects.
This ability to anticipate the placement of new elements ensures that edits are seamlessly integrated with the existing video content, leading to more natural and consistent results.

\section{More Results}
\begin{figure*}
  \centering
   \includegraphics[width=.96\linewidth]{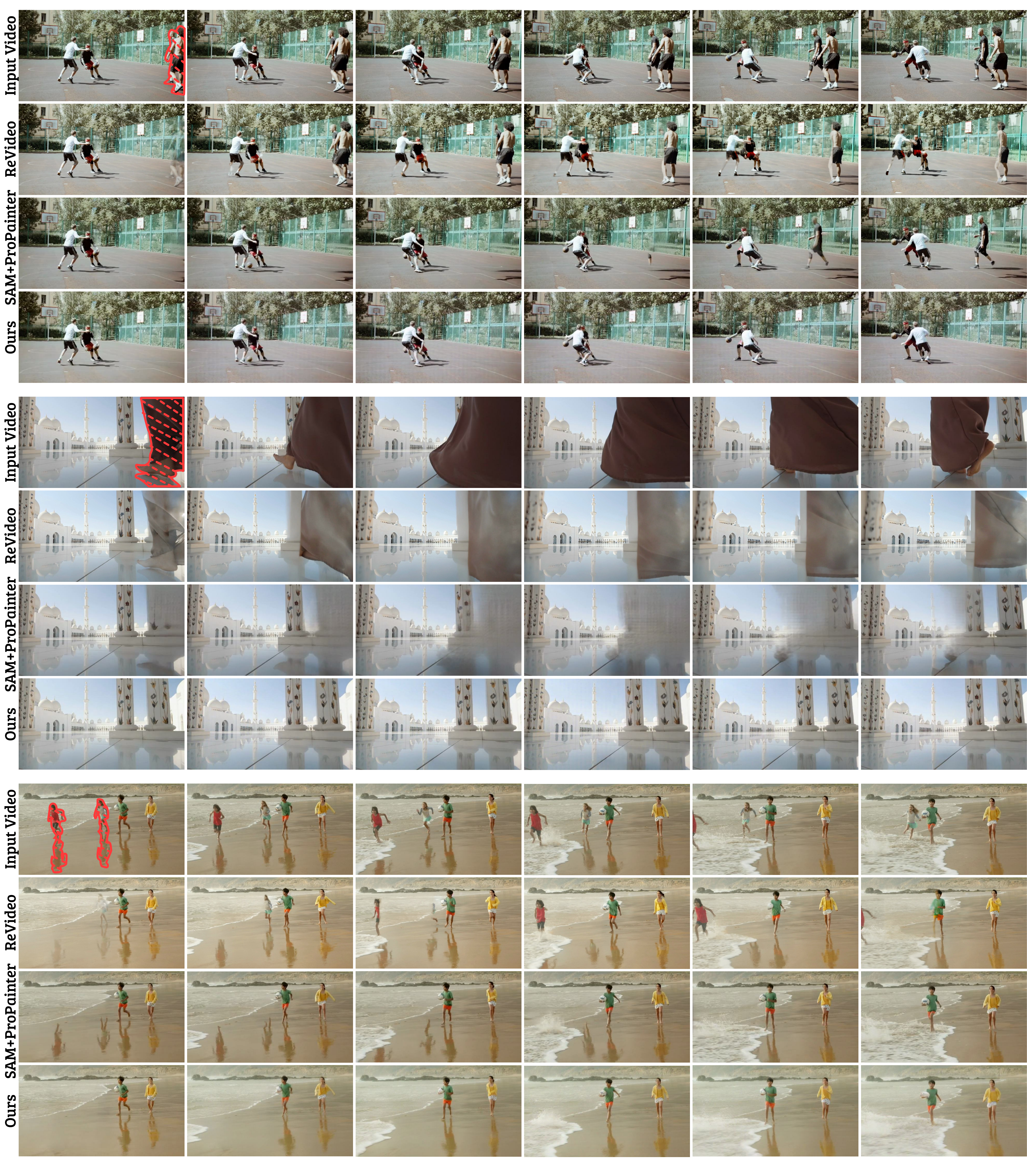}
   \caption{Additional Comparison for Removal. Our model is able to consistently remove the object and its effect (e.g., shadow, reflection) together in the whole video.}
   \vspace{-1mm}
   \label{fig:more-remove}
\end{figure*}

\begin{figure*}
  \centering
   \includegraphics[width=.96\linewidth]{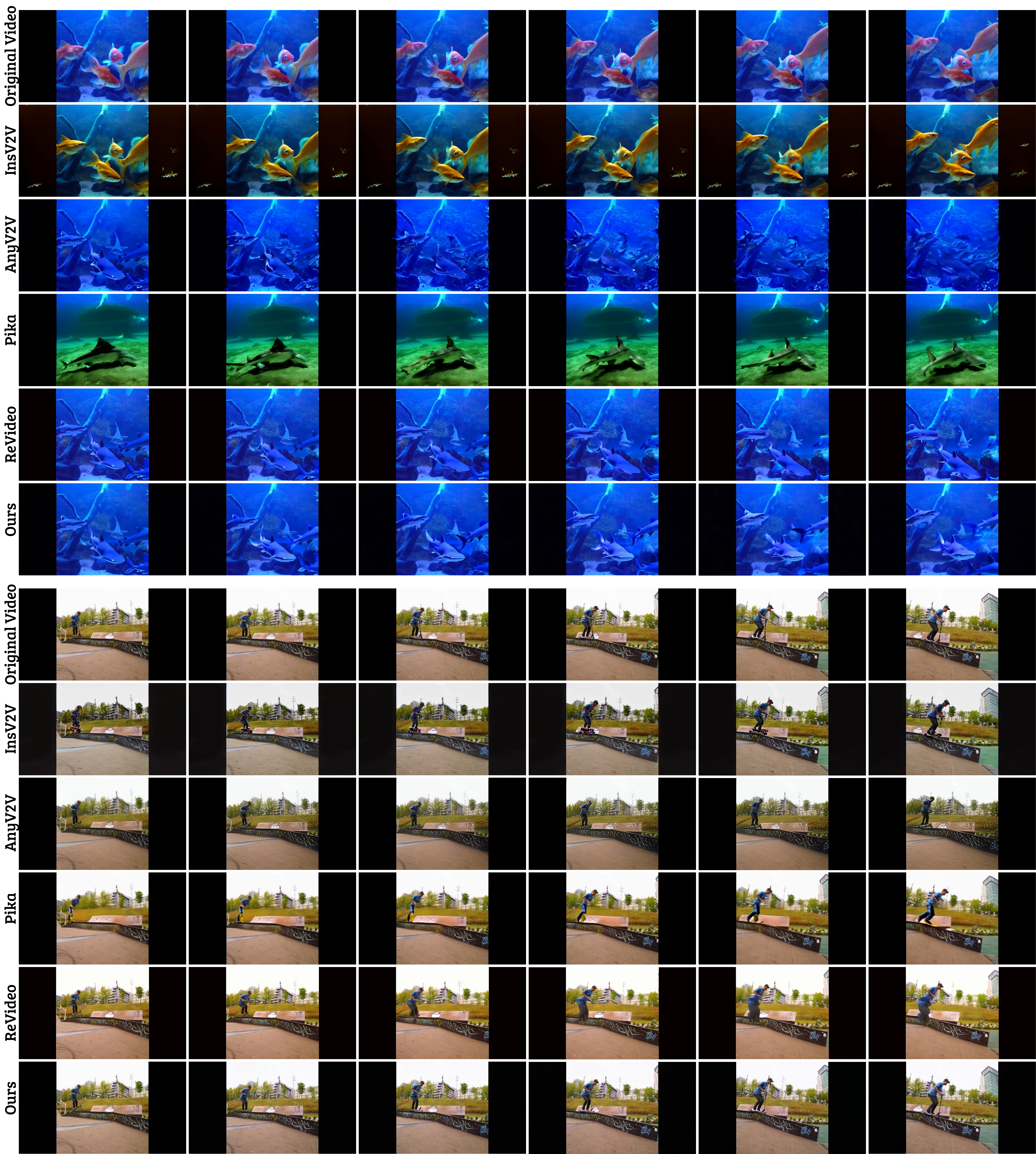}
   \caption{Additional Comparison for Editing on TGVE~\cite{wu2023cvpr}. We provide additional comparisons on the TGVE dataset~\cite{wu2023cvpr}. The first frame shown in Ours is the edited frame. As shown, our model is able to propagate the desired edit throughout the video.}
   \vspace{-1mm}
   \label{fig:more-edit-tgve}
\end{figure*}

More comparison results are shown in Fig.~\ref{fig:more-remove} (removal), Fig.~\ref{fig:more-edit-tgve} (TGVE~\cite{wu2023cvpr}), and Fig.~\ref{fig:more-edit-complex} (Challenging Test Set).
We further provide video results as part of the Supplementary Material. Please refer to the folders \texttt{1-Showcase} for various video results of our model and \texttt{2-Comparison} for video comparisons to existing work. HTML file provided inside each folder will visualize an HTML gallery with all video clips. Additionally, a demo video \texttt{demo.mp4} is provided for reference.

\begin{figure*}
  \centering
   \includegraphics[width=.96\linewidth]{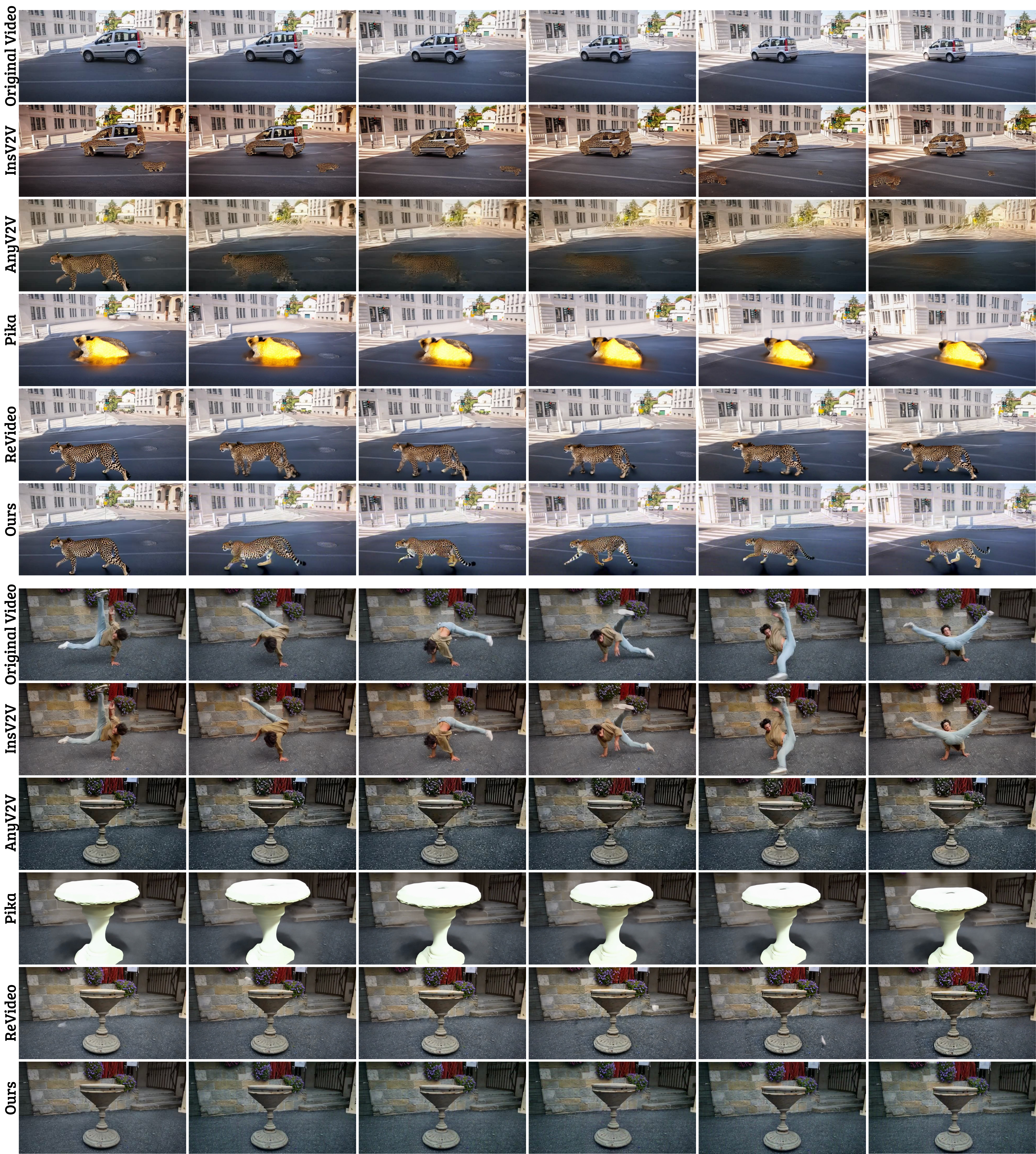}
   \caption{Additional Comparison for Editing on the Challenging Test Set. We provide additional comparisons on the Challenging Test Set. The first frame shown in Ours is the edited frame. Our model is able to replace existing objects and generate independent motion for inserted objects over the video frames.}
   \vspace{-1mm}
   \label{fig:more-edit-complex}
\end{figure*}
\clearpage

\begin{figure}
  \centering
   \includegraphics[width=.9\linewidth]{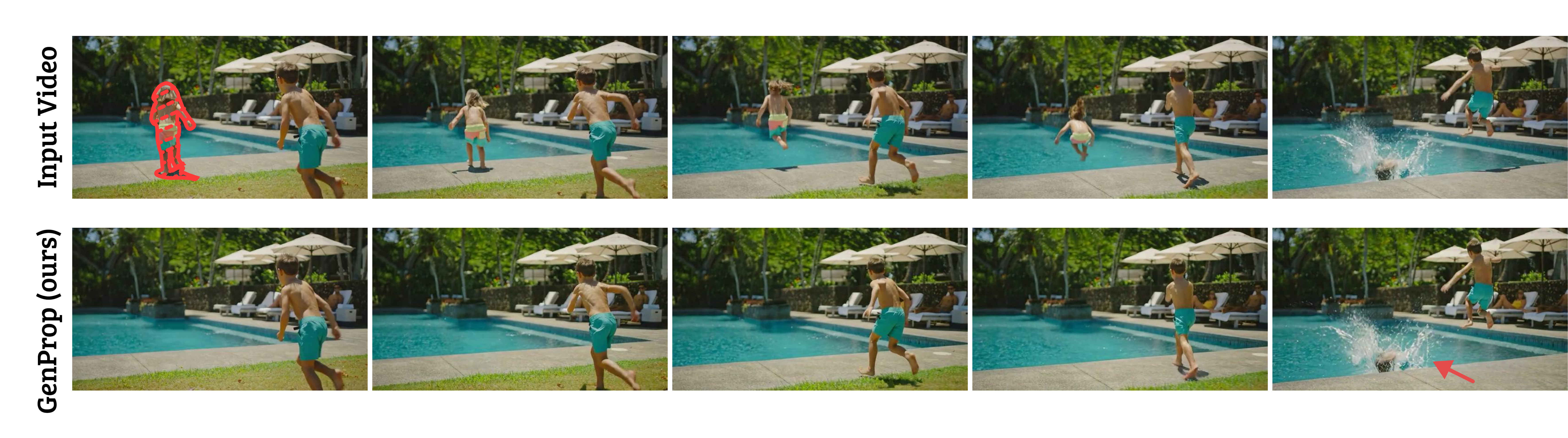}

   \caption{Limitation. It is still challenging to remove the events caused by the object, e.g., the splash of water is not removed when the girl jumping into the pool is removed.}
   \vspace{-1mm}
   \label{fig:limit}
\end{figure}

\section{Limitations}
As shown in Fig.~\ref{fig:limit}, while GenProp demonstrates the ability to handle side effects such as shadows and reflections during tasks like removal and tracking, higher-level effects caused by objects or events remain challenging to edit. For example, the splash of water generated when the girl jumps into the pool (Fig.~\ref{fig:limit}) cannot be directly modified or controlled within the current framework. This limitation presents an interesting direction for future research.

\end{document}